%% file: main.tex
\pdfoutput=1
\RequirePackage{xcolor}
\RequirePackage{fontawesome5}
\newcommand{\websiteurl}{https://rogy-labs.github.io/tutorduel/}
\newcommand{\codeurl}{https://github.com/RoGy-Labs/tutorduel-code}
\newcommand{\dataurl}{https://huggingface.co/datasets/romainpuech/tutorduel}
\newcommand{\linkrow}{%
  \href{\websiteurl}{\faGlobe\enspace\textcolor{blue}{Website}}%
  \hspace{1.6em}$|$\hspace{1.6em}
  \href{\codeurl}{\faGithub\enspace\textcolor{blue}{Code}}%
  \hspace{1.6em}$|$\hspace{1.6em}
  \href{\dataurl}{\raisebox{-0.3em}{\includegraphics[height=1.4em]{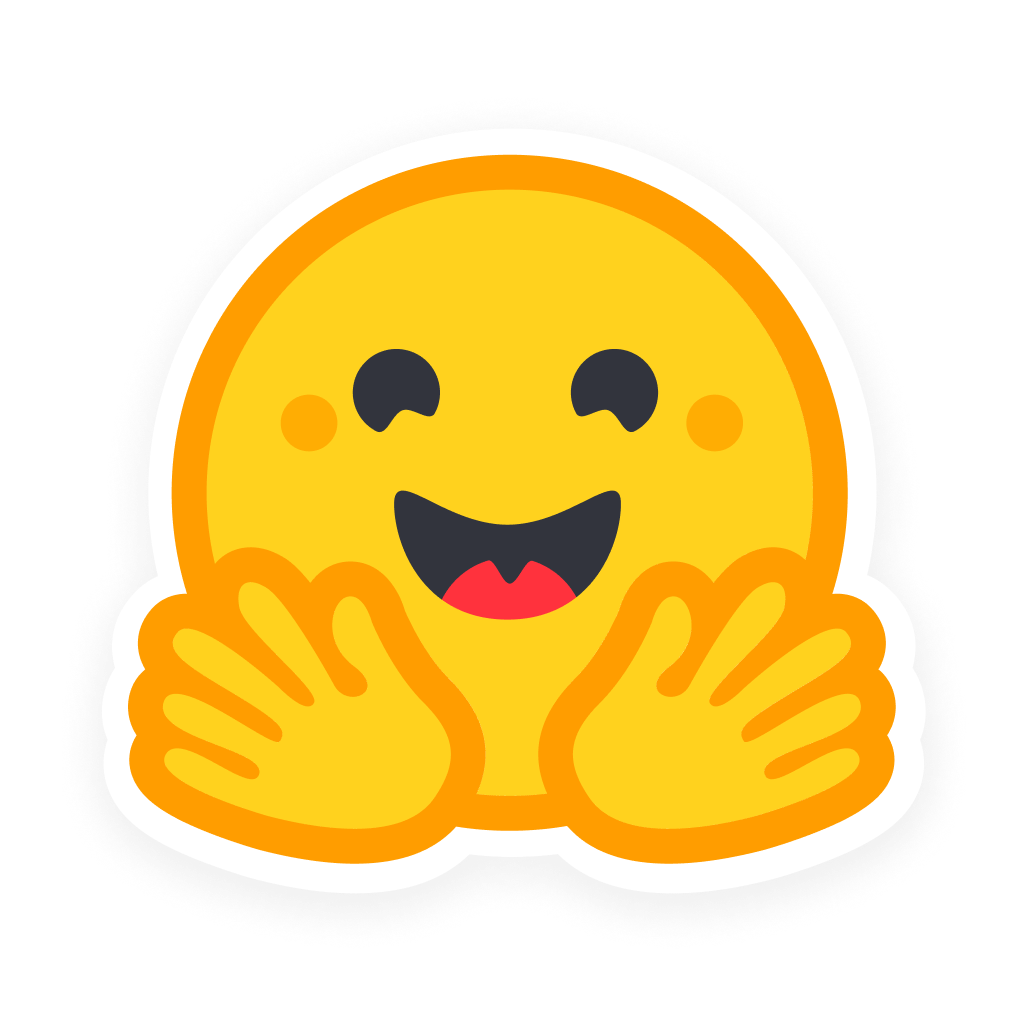}}\enspace\textcolor{blue}{Data}}%
}
\makeatletter
\AddToHook{begindocument/before}{%
  \iclrfinalcopy
  \pagestyle{plain}
  \author{%
    \begin{tabular}[t]{@{}l@{}}
      \textbf{Romain Puech}\thanks{Equal contribution.}\kern0.5em\\
      Massachusetts Institute\\
      of Technology\\
      \texttt{puech@mit.edu}
    \end{tabular}\hfill
    \begin{tabular}[t]{@{}l@{}}
      \textbf{Georgy Salakhutdinov}\footnotemark[1]\kern0.5em\\
      Massachusetts Institute\\
      of Technology\\
      \texttt{salakh@mit.edu}
    \end{tabular}\hfill
    \begin{tabular}[t]{@{}l@{}}
      \textbf{Dimitris Bertsimas}\\
      Massachusetts Institute\\
      of Technology\\
      \texttt{dbertsim@mit.edu}
    \end{tabular}%
  }
  \hypersetup{%
    pdftitle={Online LLM Personalization from Pairwise Preferences Across Users},
    pdfauthor={Romain Puech, Georgy Salakhutdinov, Dimitris Bertsimas},
    hidelinks
  }
  \renewcommand{\@maketitle}{%
    {\LARGE\scshape\raggedright\@title\par}
    \vskip 1.5em
    {\normalsize\noindent\@author\par}
    \vskip 1.2em
    {\centering\normalsize\linkrow\par}
    \vskip 0.3in minus 0.1in
  }
}
\makeatother
\documentclass{article} %
\usepackage{iclr2027_conference,times}

\usepackage{float}
\input{math_commands.tex}

\usepackage{amssymb}
\usepackage{algorithm}
\usepackage{algorithmic}
\usepackage{graphicx}
\usepackage{booktabs}
\input{prompt_boxes}

\usepackage{hyperref}
\usepackage{url}

\newcommand{\method}{CODuel}

\newcommand{\benchmark}{TutorDuel}
\input{generated/numbers}

\input{generated/agreement}

\input{generated/ceiling}
\input{generated/profiles}

\input{generated/second_judge}

\title{Online LLM Personalization from Pairwise Preferences Across Users}

\begin{document}

\maketitle

\begin{abstract}
Large language models can answer the same question in different styles, but users' preferred styles are hidden and vary across users and questions. We study inference-time personalization through online selection from a fixed set of prompts, each representing a persona. Users arrive sequentially, each asks a short sequence of questions, and the system displays two responses per question and observes one pairwise choice. We formulate this interaction as a variant of the contextual dueling bandit problem in which the learner starts without preference data, adapts within each session, and exploits information shared across users. The objective is to minimize cumulative strong regret, summing the utility gaps of both displayed responses relative to the user's best response among the available options. We make two contributions. First, we introduce the problem and release a reproducible preference benchmark and dataset built around a tutoring case study, containing educational questions, LLM-generated answers in six teaching styles, and pairwise judgments from LLM-simulated students with hidden preference profiles. We fit Bradley-Terry models to these judgments to obtain reference utilities for regret evaluation. Second, we propose a baseline that learns a mixture prior over users from completed sessions and adapts to each new user within the session. On students and topics never used in development, it removes \numTabMainMethodGapLo--\numTabMainMethodGapHi\% of random pairs' excess regret and achieves over half the performance of a per-user linear oracle fitted in hindsight. We release code, prompts, and the benchmark.
\end{abstract}

\begin{figure}[H]
\centering
\includegraphics[width=\textwidth]{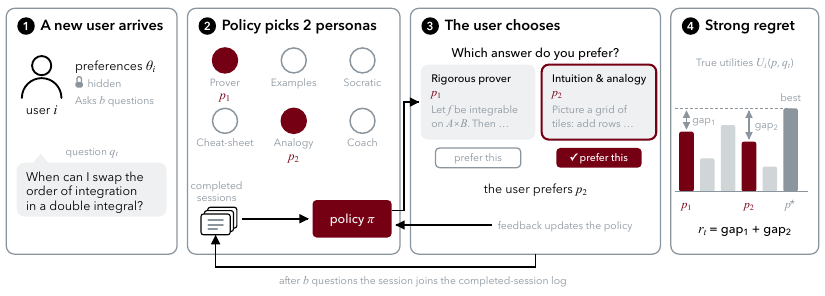}
\caption{Overview. A new user arrives with hidden preferences and asks $b$ questions. On each question, the policy picks two of the $K=6$ personas, using earlier users' sessions and the current user's choices so far, and the user chooses between the two generated answers. Strong regret, computed from hidden utilities, sums the gaps of both displayed answers to the user's best answer.}
\label{fig:overview}
\end{figure}

\input{sections/intro}
\input{sections/related}
\input{sections/setup}
\input{sections/experiments}
\input{sections/algorithm}
\input{sections/results}
\input{sections/limitations}
\input{sections/statements}

\bibliography{references}
\bibliographystyle{iclr2027_conference}

\newpage
\appendix
\input{sections/appendix_simulator}
\input{sections/appendix_uncertainty}
\clearpage
\input{sections/appendix_dataset}

\clearpage
\input{sections/appendix_prompts}
\clearpage
\input{sections/appendix_results}

\clearpage
\input{sections/appendix_ablations}

\end{document}

%% file: math_commands.tex
\usepackage{amsmath,amsfonts,bm}

\def\eqref#1{equation~\ref{#1}}

\def\1{\bm{1}}

\DeclareMathAlphabet{\mathsfit}{\encodingdefault}{\sfdefault}{m}{sl}
\SetMathAlphabet{\mathsfit}{bold}{\encodingdefault}{\sfdefault}{bx}{n}

%% file: prompt_boxes.tex
\usepackage[breakable]{tcolorbox}
\tcbset{frametitle/.style={title={#1},title after break={#1 (continued)}}}
\newtcolorbox{promptbox}[1][]{
  breakable,
  lines before break=4,
  colback=black!3,
  colframe=black!30,
  colbacktitle=black!8,
  coltitle=black,
  boxrule=0.4pt,
  arc=0pt,
  left=6pt,right=6pt,top=5pt,bottom=5pt,
  toptitle=3pt,bottomtitle=3pt,
  before skip=6pt,after skip=6pt,
  pad at break*=3pt,
  fontupper=\small\sffamily\raggedright,
  fonttitle=\small\sffamily\bfseries\raggedright,
  #1
}

%% file: generated/numbers.tex
\newcommand{\numAblAlphaFive}{+22.1}
\newcommand{\numAblAlphaFiveHi}{+40.3}
\newcommand{\numAblAlphaFiveLo}{+3.8}

\newcommand{\numAblGreedy}{+5.2}
\newcommand{\numAblGreedyHi}{+18.2}
\newcommand{\numAblGreedyLo}{-7.9}

\newcommand{\numAblGreedyShareNeg}{-0.5\%}

\newcommand{\numAblHGreedyCShareAbs}{1.4\%}

\newcommand{\numAblHGreedyUShareAbs}{3.9\%}

\newcommand{\numAblHNoGateCShareAbs}{1.5\%}

\newcommand{\numAblHNoPopCShareAbs}{28.9\%}

\newcommand{\numAblHNoPopIShareAbs}{30.7\%}
\newcommand{\numAblHNoPopRangeHi}{31}
\newcommand{\numAblHNoPopRangeLo}{29}

\newcommand{\numAblHNoPopUShareAbs}{30.9\%}

\newcommand{\numAblHNoRecalCShareAbs}{2.3\%}

\newcommand{\numAblHNoRecalIShareAbs}{4.1\%}
\newcommand{\numAblHNoRecalRangeHi}{4.1}
\newcommand{\numAblHNoRecalRangeLo}{2.3}

\newcommand{\numAblHNoRecalUShareAbs}{2.8\%}

\newcommand{\numAblHOneCompCShareAbs}{9.4\%}

\newcommand{\numAblHOneCompIShareAbs}{11.8\%}
\newcommand{\numAblHOneCompRangeHi}{30}
\newcommand{\numAblHOneCompRangeLo}{9}

\newcommand{\numAblHOneCompUShareAbs}{30.0\%}

\newcommand{\numAblHOneGaussCShareAbs}{2.2\%}

\newcommand{\numAblHOneGaussIShareAbs}{2.4\%}

\newcommand{\numAblHOneGaussUShareAbs}{7.0\%}
\newcommand{\numAblHSeedsC}{32}

\newcommand{\numAblLts}{+213.0}

\newcommand{\numAblLucb}{+128.5}

\newcommand{\numAblMoment}{+3.7}

\newcommand{\numAblNoGate}{+8.1}

\newcommand{\numAblNoPop}{+386.2}

\newcommand{\numAblNoRecal}{+41.2}
\newcommand{\numAblNoRecalHi}{+52.2}
\newcommand{\numAblNoRecalLo}{+30.2}

\newcommand{\numAblNoTransfer}{+449.7}

\newcommand{\numAblOneComp}{+157.7}

\newcommand{\numAblOneCompShareAbs}{16.2\%}

\newcommand{\numAblOneGauss}{-2.5}
\newcommand{\numAblOneGaussHi}{+9.3}
\newcommand{\numAblOneGaussLo}{-14.3}
\newcommand{\numAblOneGaussShare}{-0.3\%}

\newcommand{\numAblSeeds}{32}
\newcommand{\numCBestOtherGap}{38}

\newcommand{\numCClosestShareAbs}{9.3\%}

\newcommand{\numCIRsOverClosest}{7}

\newcommand{\numCLinGap}{81}

\newcommand{\numCMethodGap}{44}

\newcommand{\numCOracleBestScore}{39}

\newcommand{\numCOracleMethodScore}{44}

\newcommand{\numCPerUserGap}{66}

\newcommand{\numCRsMethodGap}{45}
\newcommand{\numCRsMethodGapHi}{47}
\newcommand{\numCRsMethodGapLo}{42}

\newcommand{\numCRsOverColdb}{+6}

\newcommand{\numCShareAbs}{23.1\%}

\newcommand{\numColdbLongClusters}{2.0}

\newcommand{\numColdbLongPooled}{2489}
\newcommand{\numColdbLongSplit}{865}
\newcommand{\numColdbLrShortClusters}{3.2}

\newcommand{\numDataCVarType}{30.4}

\newcommand{\numDataDVarQuestion}{48.3}

\newcommand{\numDataDVarType}{31.7}

\newcommand{\numDataIVarType}{34.5}
\newcommand{\numDataProfSimMax}{0.96}
\newcommand{\numDataProfiles}{16}

\newcommand{\numDataUVarType}{50.3}

\newcommand{\numDevGridRestHi}{34}
\newcommand{\numDevGridRestLo}{28}
\newcommand{\numDevGridShareHi}{+34}
\newcommand{\numDevGridShareLo}{+6}

\newcommand{\numEarlyGainHi}{16}
\newcommand{\numEarlyGainLo}{14}
\newcommand{\numGateScrambledGatedShare}{+0.9\%}

\newcommand{\numGateScrambledUngatedShare}{-0.7\%}

\newcommand{\numGateSimNullGatedShare}{-0.3\%}

\newcommand{\numGateSimNullUngatedShare}{-12.6\%}
\newcommand{\numGateSimNullUngatedShareAbs}{13\%}
\newcommand{\numIBestOtherGap}{38}

\newcommand{\numIClosestShareAbs}{12.5\%}

\newcommand{\numILinGap}{82}

\newcommand{\numIMethodGap}{46}

\newcommand{\numIOracleBestScore}{40}

\newcommand{\numIOracleMethodScore}{46}

\newcommand{\numIPerUserGap}{68}

\newcommand{\numIRsMethodGap}{46}
\newcommand{\numIRsMethodGapHi}{49}
\newcommand{\numIRsMethodGapLo}{43}

\newcommand{\numIRsOverColdb}{+8}

\newcommand{\numIShareAbs}{25.2\%}

\newcommand{\numJVAllFourSameSignHi}{55}
\newcommand{\numJVAllFourSameSignLo}{54}
\newcommand{\numJVLeanFirstShownHi}{0.48}
\newcommand{\numJVLeanFirstShownLo}{0.24}
\newcommand{\numJVLeanLabelIAHi}{0.23}
\newcommand{\numJVLeanLabelIALo}{0.13}
\newcommand{\numJVLenEffectSdHi}{-0.090}
\newcommand{\numJVLenEffectSdLo}{-0.161}
\newcommand{\numJVLongerWinsHi}{52}
\newcommand{\numJVLongerWinsLo}{44}
\newcommand{\numJVPairsSaturatedHi}{41}
\newcommand{\numJVPairsSaturatedLo}{38}
\newcommand{\numJVPrompts}{2{,}448{,}000}
\newcommand{\numJVPromptsAtCapHi}{79}
\newcommand{\numJVPromptsAtCapLo}{73}

\newcommand{\numJVRLabelHalvesHi}{0.90}
\newcommand{\numJVRLabelHalvesLo}{0.89}
\newcommand{\numJVROrderHalvesHi}{0.72}
\newcommand{\numJVROrderHalvesLo}{0.65}
\newcommand{\numJVRelPicksHi}{0.99}
\newcommand{\numJVRelPicksLo}{0.99}
\newcommand{\numJVRelQuestionDevHi}{0.87}
\newcommand{\numJVRelQuestionDevLo}{0.82}
\newcommand{\numJVRelStudentHi}{0.96}
\newcommand{\numJVRelStudentLo}{0.94}

\newcommand{\numJVWordsConcise}{107}
\newcommand{\numJVWordsOtherHi}{169}
\newcommand{\numJVWordsOtherLo}{147}
\newcommand{\numLateGainHi}{31}
\newcommand{\numLateGainLo}{27}
\newcommand{\numLibGreedy}{961.8}
\newcommand{\numLibLookahead}{954.6}
\newcommand{\numLibLts}{1197.6}
\newcommand{\numLibLucb}{1114.6}
\newcommand{\numNewStudentsMethodDropHi}{10}
\newcommand{\numNewStudentsMethodDropLo}{6}

\newcommand{\numScrambledBetter}{3}

\newcommand{\numScrambledGridHi}{+1}
\newcommand{\numScrambledGridLo}{-1}

\newcommand{\numSeedsC}{32}
\newcommand{\numSeedsD}{32}

\newcommand{\numSeedsT}{64}
\newcommand{\numSeedsU}{64}

\newcommand{\numSensBelowClosest}{42}
\newcommand{\numSensBest}{+0.9}
\newcommand{\numSensCells}{42}

\newcommand{\numSensSeedsC}{32}

\newcommand{\numSensSigBelowClosest}{42}
\newcommand{\numSensWorst}{-1.0}

\newcommand{\numShareAbsMinLo}{21.1}

\newcommand{\numSimAnchorCells}{16}

\newcommand{\numSimGridBetter}{155}
\newcommand{\numSimGridCells}{160}

\newcommand{\numSimNullCells}{20}

\newcommand{\numSimNullMeanShare}{-0.1}

\newcommand{\numSimNullSharersHi}{207\%}
\newcommand{\numSimNullSharersLo}{48\%}
\newcommand{\numSimNullWorse}{1}
\newcommand{\numSimRemovedEightTypesHi}{64}
\newcommand{\numSimRemovedEightTypesLo}{33}

\newcommand{\numSimRemovedGaussHi}{28}
\newcommand{\numSimRemovedGaussLo}{24}

\newcommand{\numSimRemovedTwoTypesHi}{85}
\newcommand{\numSimRemovedTwoTypesLo}{76}
\newcommand{\numSimTrendBFirst}{3}
\newcommand{\numSimTrendBHi}{+53}
\newcommand{\numSimTrendBLast}{48}
\newcommand{\numSimTrendBLo}{+39}
\newcommand{\numSimTrendBetaFirst}{1}
\newcommand{\numSimTrendBetaHi}{+56}
\newcommand{\numSimTrendBetaLast}{8}
\newcommand{\numSimTrendBetaLo}{+36}

\newcommand{\numTMethodGap}{54}

\newcommand{\numTTypeLibGap}{54}

\newcommand{\numTabMainMethodGapHi}{48}
\newcommand{\numTabMainMethodGapLo}{44}
\newcommand{\numTabMainMethodOfLinHi}{57}
\newcommand{\numTabMainMethodOfLinLo}{54}

\newcommand{\numTsPairMaxPct}{2.1}

\newcommand{\numTuneMetaTsConfigs}{63}
\newcommand{\numTypeLibMainHi}{33}
\newcommand{\numTypeLibMainLo}{23}

\newcommand{\numUBestOtherGap}{33}

\newcommand{\numUClosestShareAbs}{22.5\%}

\newcommand{\numULinGap}{85}

\newcommand{\numUMetaTsGap}{23}

\newcommand{\numUMethodGap}{48}

\newcommand{\numUMethodProfileBest}{61}
\newcommand{\numUMethodProfileWorst}{29}

\newcommand{\numUNoTransferGap}{30}

\newcommand{\numUOracleBestScore}{34}
\newcommand{\numUOracleColdbBest}{32}
\newcommand{\numUOracleColdbPaper}{22}

\newcommand{\numUOracleMethodScore}{48}

\newcommand{\numUPerUserGap}{76}

\newcommand{\numUProfileBootHi}{55}
\newcommand{\numUProfileBootLo}{40}
\newcommand{\numUProfileOtherName}{CoLSTIM}

\newcommand{\numURsDistinct}{5.2}

\newcommand{\numURsMethodGap}{55}

\newcommand{\numURsSeeds}{64}

\newcommand{\numUShareAbs}{25.9\%}

%% file: generated/agreement.tex
\newcommand{\numAgrAuc}{0.79}
\newcommand{\numAgrAucHi}{0.84}

\newcommand{\numAgrAucLo}{0.73}
\newcommand{\numAgrBinCapExp}{82}
\newcommand{\numAgrBinCapRater}{82}
\newcommand{\numAgrBinLowExp}{52}

\newcommand{\numAgrBinLowRater}{55}
\newcommand{\numAgrBinMidHit}{66}
\newcommand{\numAgrBinMidP}{0.80}
\newcommand{\numAgrBinTopHit}{81}

\newcommand{\numAgrBinTopP}{0.8}
\newcommand{\numAgrBinTopShare}{63}
\newcommand{\numAgrBothAgreeJudge}{119}

\newcommand{\numAgrBothAgreeN}{142}
\newcommand{\numAgrCells}{51}

\newcommand{\numAgrDiff}{+3.9}

\newcommand{\numAgrDiffHi}{+10.0}

\newcommand{\numAgrDiffLo}{-2.5}
\newcommand{\numAgrDiffOpus}{+0.5}
\newcommand{\numAgrDiffOpusHi}{+4.9}
\newcommand{\numAgrDiffOpusLo}{-3.9}
\newcommand{\numAgrDiffSol}{+1.0}
\newcommand{\numAgrDiffSolHi}{+3.9}
\newcommand{\numAgrDiffSolLo}{-2.0}
\newcommand{\numAgrEce}{8}

\newcommand{\numAgrJevAgree}{73.5}

\newcommand{\numAgrJudgeAgree}{73.5}

\newcommand{\numAgrJudgeRone}{74.0}
\newcommand{\numAgrJudgeRtwo}{73.0}
\newcommand{\numAgrJudgesAgree}{81.4}

\newcommand{\numAgrKappaJudgeRone}{0.48}
\newcommand{\numAgrKappaJudgeRtwo}{0.46}

\newcommand{\numAgrKappaRaters}{0.39}
\newcommand{\numAgrLabels}{408}

\newcommand{\numAgrPairs}{204}

\newcommand{\numAgrRaterAgree}{69.6}

\newcommand{\numAgrSameFavHighOpus}{86.8}

\newcommand{\numAgrSameFavHighSol}{93.8}

\newcommand{\numAgrSamplePairs}{300}
\newcommand{\numAgrSampleUnits}{75}

%% file: generated/profiles.tex
\newcommand{\numProfFavDevseven}{78}

\newcommand{\numProfFavNewfour}{97}

\newcommand{\numProfFavNewseven}{87}

%% file: generated/second_judge.tex
\newcommand{\numSJAgree}{80.7}

\newcommand{\numSJAgreeComparisons}{378{,}000}

\newcommand{\numSJAgreeTop}{93.8}
\newcommand{\numSJAgreeTopShare}{39}

\newcommand{\numSJClickCross}{0.66}
\newcommand{\numSJClickSelf}{0.70}
\newcommand{\numSJClickSelfJ}{0.72}
\newcommand{\numSJStudents}{1{,}050}
\newcommand{\numSJTopPersona}{79}

\newcommand{\numSJTopPersonaChance}{17}

\newcommand{\numSJTransferC}{73}
\newcommand{\numSJTransferI}{76}
\newcommand{\numSJTransferOwnC}{83}
\newcommand{\numSJTransferOwnI}{83}
\newcommand{\numSJTransferOwnU}{87}
\newcommand{\numSJTransferU}{85}
\newcommand{\numJCBestOtherGap}{44}

\newcommand{\numJCClosestShareAbs}{20.4\%}

\newcommand{\numJCMethodGap}{55}

\newcommand{\numJDMethodGap}{59}

\newcommand{\numJIBestOtherGap}{40}

\newcommand{\numJIClosestShareAbs}{21.8\%}

\newcommand{\numJIMethodGap}{53}

\newcommand{\numJLinRange}{90}
\newcommand{\numJMethodOfLinHi}{65}
\newcommand{\numJMethodOfLinLo}{59}
\newcommand{\numJSeedsC}{32}
\newcommand{\numJSeedsD}{32}

\newcommand{\numJSeedsT}{64}
\newcommand{\numJSeedsU}{64}

\newcommand{\numJTMethodGap}{62}

\newcommand{\numJUBestOtherGap}{45}

\newcommand{\numJUClosestShareAbs}{25.4\%}

\newcommand{\numJUMethodGap}{59}

%% file: sections/intro.tex
\section{Introduction}
\label{sec:intro}

A large language model (LLM) chatbot can answer the same question in several styles, and users differ in the style they prefer.
A common way to guide the response style is to use a system prompt that specifies a \emph{persona}~\citep{zheng2024helpful}.
For example, an LLM math tutor can adopt a persona that emphasizes proofs, worked examples, or guiding questions.
The preferred persona depends on the user and the question: the same user may want a proof for a definition and numbered steps for a computation.
A simple way to collect preference feedback is to show two answers and ask the user to choose one.

In our setup, users arrive one after another, each for a short session, and the platform starts with no preference data.
On every question it selects two personas using features of the persona-question pairs and the choices observed so far.
A user's preferences are hidden and fixed during the session, and later users can benefit only from the log the policy itself created.
Shared structure across the population is what makes that transfer possible.

We study inference-time personalization through online selection from this fixed catalog of prompts. We make two contributions.

\begin{enumerate}
\item \textbf{Problem formulation and dataset.}
We formalize the interaction as a multi-user variant of the contextual dueling bandit problem, with cumulative strong regret accounting for the utility gaps of both displayed answers~\citep{yue2012karmed,bengs2022stochastic,wang2025online}.
We introduce \textbf{\benchmark{}}, a reproducible tutoring preference benchmark and dataset containing educational questions, LLM-generated answers in six teaching styles, and pairwise judgments from LLM-simulated students with hidden preference profiles.
On comparisons labeled by two of the authors for the described students, the judge agrees with each author at least as often as the two authors agree with each other (\numAgrJudgeRone\% and \numAgrJudgeRtwo\% against \numAgrRaterAgree\%). We provide reference utilities for regret evaluation by fitting Bradley-Terry models to the LLM judge's choice probabilities. The evaluation protocol brackets performance between picking personas at random and per-user references, and studies the generalization to new topics and new students never used in development.
We also provide a numerical simulator for controlled evaluation.

\item \textbf{A baseline algorithm.}
We propose \textbf{\method{}} as a baseline for this problem. It learns a mixture prior over user preferences from completed sessions to starts each new user from that prior.
On three \benchmark{} rosters whose topics and students were never used in design (eight new hand-written profiles, and two rosters of $350$ distinct students composed from preference traits), it removes \numUMethodGap\%, \numCMethodGap\%, and \numIMethodGap\% of the excess regret of random pairs (Table~\ref{tab:main}). This is \numTabMainMethodOfLinLo\% to \numTabMainMethodOfLinHi\% of the performance of a per-user linear oracle, which chooses each user's parameters in hindsight on the features of \method{}; we offer the difference as a target for future methods.
\end{enumerate}

%% file: sections/related.tex
\section{Related Work}
\label{sec:related}

\textbf{Contextual dueling bandits.}
Contextual dueling bandits extend dueling bandits~\citep{yue2012karmed,bengs2021preference} to arms that depend on a context, as in CoLSTIM~\citep{bengs2022stochastic}, the algorithms of \citet{saha2021optimal}, and feel-good Thompson sampling~\citep{li2024feelgood}.
Related RLHF formulations treat prompts as contexts and responses as arms~\citep{zhu2023principled,xiong2024iterative,das2024active}.
COLDB~\citep{wang2025online} extends clustering of bandits~\citep{gentile2014online,li2016collaborative} to dueling feedback, pooling observations within clusters of recurring users. Our users instead arrive once, and their preferences need not form discrete clusters.
Our pooled RUCB baseline~\citep{zoghi2014relative} ignores the question.

\textbf{Learning priors across tasks.}
Meta-Thompson sampling and AdaTS learn priors across bandit tasks from rewards~\citep{kveton2021meta,basu2021no}. Related work studies transfer across linear and hierarchical bandit tasks~\citep{cella2020meta,peleg2022metalearning,hong2022hierarchical}. MixTS uses a given mixture prior~\citep{hong2022thompson}, and \citet{simchowitz2021bayesian} analyze sensitivity to prior misspecification. \method{} learns a mixture prior from completed sessions' pairwise choices.

\textbf{Latent bandits.}
Latent bandits infer a hidden user type from feedback while minimizing regret~\citep{maillard2014latent}. \citet{hong2020latent} develop Thompson sampling and UCB rules under uncertain type-specific reward models. \method{} similarly keeps a belief over mixture components within each session, with pairwise feedback and regret on both displayed answers.

\textbf{Personalization across users.}
Personalized alignment uses policy merging~\citep{jang2023personalized}, user embeddings~\citep{li2024personalized,poddar2024personalizing}, personalized reward models~\citep{park2024rlhf}, preference prototypes~\citep{chen2025pal}, low-rank reward bases~\citep{bose2025lore}, and meta-learned initializations~\citep{cai2026meta}. Gaussian-process preference models share information across users and actively select comparisons~\citep{bonilla2010gaussian,houlsby2012collaborative}. Our protocol starts without preference data and measures strong regret on every displayed pair across sequential short user sessions.
Online per-user methods are closer in protocol. UserAlign~\citep{padurean2025inferencetime} selects a response for one user and prompt from a fixed pool by best-arm identification in a logistic bandit, and T-POP~\citep{qu2025tpop} steers decoding with a reward learned by a neural dueling bandit. Both learn each user from scratch, like our CoLSTIM baseline. UserAlign reports final-response win rate and query count, rather than cumulative regret on displayed pairs.

\textbf{Simulated users and personalization benchmarks.}
Prior work uses LLMs to simulate human responses~\citep{argyle2023out,aher2023using}. These simulations condition on personality profiles~\citep{jiang2024personallm}, personal interviews~\citep{park2024generative}, and student knowledge and personality~\citep{lu2024generative,liu2024personality}. AlpacaFarm~\citep{dubois2023alpacafarm}, PERSONA~\citep{castricato2025persona}, and PersonalLLM~\citep{zollo2024personalllm} provide simulated preference data. PRISM collects human preferences~\citep{kirk2024prism}, while LaMP evaluates personalization from user histories~\citep{salemi2024lamp}. \benchmark{} builds on this use of simulation by prompting an LLM with a student profile to judge tutor answers. Such simulations can misrepresent group opinions and response variability~\citep{santurkar2023whose,bisbee2024synthetic} (Section~\ref{sec:limitations}).

%% file: sections/setup.tex
\section{Problem Setting}
\label{sec:setup}

We consider users who interact with an LLM chatbot by asking a sequence of questions. For each question, the chatbot presents two alternative answers, and the user chooses the one they prefer.

We write $\lceil z\rceil$ for the smallest integer at least $z$, and vectors without boldface.

\textbf{Users and population.}
A chatbot serves $u\geq 1$ users sequentially, indexed by $\mathcal{U}=[u]=\{1,\ldots,u\}$. Each user $i\in\mathcal{U}$ arrives once, completes one uninterrupted session of $b\geq 1$ questions, and leaves. Time is global: $t\in[T]$ with horizon $T=ub$ and current user $i_t=\lceil t/b\rceil$, so user $i$ is served at rounds $t=(i-1)b+1,\ldots,ib$.

Each user has a hidden preference vector $\theta_i\in\mathbb{R}^d$, drawn independently from a fixed, unknown distribution $\mathcal{P}$, $\theta_1,\ldots,\theta_u\overset{\mathrm{i.i.d.}}{\sim}\mathcal{P}$, with $\|\theta_i\|_2\le B_\theta$ for a known bound $B_\theta>0$. The vector $\theta_i$ stays fixed throughout the session.
Examples include finite-type populations, Gaussian clouds, and low-rank Gaussian clouds (Section~\ref{sec:exp-simulator} and Appendix~\ref{app:simulator}), and, in the tutoring benchmark, students who combine a few shared preference traits (Section~\ref{sec:exp-llm}).

\textbf{Personas, questions, and utilities.}
The chatbot has a finite catalog of $K$ personas. In a tutoring example, a persona can be a system prompt specifying a style such as proofs, worked examples, or guiding questions. Persona $p\in[K]$ on question $q\in\mathcal{Q}$ produces an answer $x(p,q)$. For a given question $q$, the available arms are the $K$ persona-question pairs $(p,q)$. The learner selects which two personas to compare on that question.

The true utility of persona $p$ on question $q$ is that of the answer $x(p,q)$ it produces, written $f_i(p,q)$. The optimal persona may be different for each question. We assume a linear utility model~\citep{saha2021optimal,bengs2022stochastic}:
\begin{equation}
\label{eq:linear}
f_i(p,q)=\theta_i^\top\phi(p,q).
\end{equation}
Here, $\phi(p,q)\in\mathbb{R}^d$ is a known vector encoding features of persona $p$ in the context of question $q$, and $\theta_i$ encodes the user's preferences over these features. The features are available from the persona specification and question before generating the answer.
Because $\phi$ depends on both persona and question, the preferred persona can vary with $q$ even though $\theta_i$ is fixed.

At round $t$ the environment supplies $q_t\in\mathcal{Q}$. Questions need not be independent or identically distributed and may be chosen adversarially based on past interactions. The current question and the $K$ vectors $\phi(p,q_t)$ are observed before the pair is chosen.

\textbf{Interaction.}
For each $t=1,\ldots,T$, the learner observes $i_t$, $q_t$, and $\phi(p,q_t)$ for all $p\in[K]$. It then selects an ordered pair $(p_{t,1},p_{t,2})$ of distinct personas and displays their answers. The user's choice is recorded as $y_t\in\{0,1\}$: $y_t=1$ if the answer from $p_{t,1}$ is chosen, and $y_t=0$ if the answer from $p_{t,2}$ is chosen. There is one comparison per question. After the choice at $t=ib$, the session ends and its record joins the completed-user log. Let $\mathcal{H}_{t-1}$ collect observations from rounds $s<t$, with $\mathcal{H}_0=\varnothing$. An admissible policy $\pi$ chooses the current pair from this history, the current question and features, the persona specifications, the known parameters $d,K,B_\theta,\beta,b,u$, and its own randomness. Preference vectors, type labels, population parameters, and utilities remain hidden. Each policy learns from the comparisons it selected.

Because $\phi(p,q)$ is computed from the persona specification and the question, only the two selected answers need to be generated.

\textbf{Preference feedback.}
We model the binary preference following the Bradley-Terry-Luce (BTL) model~\citep{bradley1952rank,luce1959individual,bengs2022stochastic}. Write $\sigma(z)=(1+e^{-z})^{-1}$ for the logistic sigmoid. Conditional on the hidden vector, history, question, and selected pair,
\begin{equation}
\label{eq:btl}
\Pr\!\bigl(y_t=1\mid\theta_{i_t},\mathcal{H}_{t-1},q_t,p_{t,1},p_{t,2}\bigr)
=\sigma\!\bigl(\beta\bigl[f_{i_t}(p_{t,1},q_t)-f_{i_t}(p_{t,2},q_t)\bigr]\bigr),
\end{equation}
where $\beta>0$ is a known parameter controlling feedback noise, shared across users and rounds. For a fixed nonzero utility gap, larger $\beta$ makes the higher-utility answer more likely to be chosen, while smaller $\beta$ moves the choice probability toward $1/2$. Under (\ref{eq:linear}), each comparison is a logistic regression observation whose features are the difference between the two candidates' feature vectors. This law compares the true utilities of the two displayed answers and holds at every round, including adaptively selected comparisons.

\textbf{Regret.}
For evaluation, utilities are normalized to $U_i(p,q)\in[0,1]$ without changing persona rankings (Section~\ref{sec:exp-llm}). Let $p_t^\star\in\arg\max_{p\in[K]}U_{i_t}(p,q_t)$ be the best persona for user $i_t$ on question $q_t$, with ties resolved by a fixed rule. We use strong regret~\citep{yue2012karmed}, summing the utility gaps of both displayed answers as in \citet{wang2025online}:
\begin{equation}
\label{eq:regret}
r_t=2U_{i_t}(p_t^\star,q_t)-U_{i_t}(p_{t,1},q_t)-U_{i_t}(p_{t,2},q_t),\qquad
R_T=\sum_{t=1}^{T}r_t.
\end{equation}
Our problem is a multi-user variant of contextual dueling bandits~\citep{saha2021optimal,bengs2022stochastic,wang2025online}: the question determines the available arms' features, and the learner observes a pairwise choice. The objective is to minimize $\mathbb{E}_\pi[R_T]$ over admissible policies, counting every round of every user. Because two distinct personas must be shown, even a policy that knows $\theta_{i_t}$ pays the gap between the best and the second-best persona at every round. Summing these gaps gives the floor $G_T=\sum_{t=1}^{T}\bigl(U_{i_t}(p_t^\star,q_t)-\max_{p\neq p_t^\star}U_{i_t}(p,q_t)\bigr)$, so $R_T\ge G_T$. We define \emph{excess strong regret} as $R_T-G_T$ and report the share of random pairs' excess regret that a policy removes (Section~\ref{sec:exp-protocol}).

%% file: sections/experiments.tex
\section{Experimental Setup}
\label{sec:experiments}

We evaluate \method{} in two environments: a numerical simulator in which the linear utility and BTL assumptions of Section~\ref{sec:setup} hold exactly, and a tutoring benchmark in which an LLM role-plays students described in natural language and compares generated tutor answers. All preferences are simulated. Pairwise choices made by two of the authors serve to validate the judge (Section~\ref{sec:exp-llm}).

\subsection{Shared Protocol and Evaluation}
\label{sec:exp-protocol}

In both environments we use $K=6$ teaching personas, and $u$ users arrive sequentially, each asking $b$ questions ($b=24$ in the tutoring benchmark). In the tutoring benchmark, we make every design choice, including the final variant of \method{}, on a development roster and run the evaluation rosters (Section~\ref{sec:exp-llm}) only after these choices are fixed. Every policy starts from the first user of each roster, with no information carried over from development, and is scored on cumulative regret of every round.

\textbf{Matched randomness.}
For each environment and seed, all policies see the same sequence of users and questions. They share one uniform draw per round to sample choices, using the BTL model~(\ref{eq:btl}) in the simulator and the stored judge probabilities in \benchmark{} (Section~\ref{sec:exp-llm}). Identical ordered pairs therefore receive identical choices.

\textbf{Benchmark score.}
In the tutoring benchmark we report the share of random pairs' excess regret that a policy removes, $100\,(R_T^{\mathrm{rand}}-R_T)/(R_T^{\mathrm{rand}}-G_T)$, where $R_T^{\mathrm{rand}}$ is the expected regret of random pairs. Random pairs score $0$, and a policy would score $100$ only by reaching the floor $G_T$ on every question. The floor follows every question-specific preference, including judge disagreement between display orders, so no policy is expected to reach it. As a reference for what a learner can reach, we report a per-user linear oracle. It uses the features of \method{}, in which the rater's score is replaced by its recalibration (Section~\ref{sec:algorithm}), with the recalibration fitted on all users of the roster in hindsight. For each user, the oracle chooses the parameters $\theta$ that minimize that user's excess regret over all their questions when it shows the two personas with the highest utilities. No policy that ranks personas with one fixed linear function of these features per user does better on that user's questions. Because the minimization has no closed form, we search over the weight on the recalibrated score with a grid and over the persona offsets by coordinate descent, and report the best value found, so the oracle's true score is at least this high. The oracle knows each user's parameters from the first round and pays no learning cost.

\subsection{Numerical Simulator}
\label{sec:exp-simulator}

The numerical simulator evaluates \method{} where the linear utility and BTL assumptions hold exactly, with known features in $d=8$ dimensions. Users are drawn from one of four populations: a few fixed types, a Gaussian cloud, a low-rank Gaussian cloud, or a null population with no shared structure, which serves as a negative control for transfer. We run $u=300$ users with $b=20$ and $\beta\in\{4,12\}$ (Appendix~\ref{app:simulator}).

\subsection{\benchmark{}: Tutoring Preference Dataset}
\label{sec:exp-llm}

\benchmark{} contains tutoring questions, generated tutor answers, pairwise choice probabilities from LLM-simulated students, and evaluation utilities. Policies receive persona-question features and sampled pairwise choices. The answer texts, probabilities, and utilities remain hidden from them.

\textbf{Personas and questions.}
The six tutor personas are a rigorous prover, a worked-examples tutor, a Socratic guide, an intuition-and-analogy tutor, a concise cheat-sheet, and a patient step-by-step coach. Tutor answers are generated with \texttt{gpt-5.6-terra} and a fixed system prompt for each persona at temperature $1$. The full tutor prompts are given in Appendix~\ref{app:tutor-prompts}.

We use \texttt{gpt-5.6-sol} to generate $24$ distinct questions per session across four topics. Each topic has two sequences covering different aspects, each with a definition question, a computation or numerical example, and a why-question. It also includes a first-person cue describing the student's situation or need (Appendix~\ref{app:cues}).

\textbf{Persona-question features.}
We compute $\phi(p,q)$ from the question text and the persona prompts alone, before any answer is generated. A rater model scores how well each persona's answer style suits the question. We use TypeSafe Jev (\texttt{jev-1.13.0})~\citep{almeida2026jev,typesafe2026docs} as the rater because it returns probabilities and is inexpensive. One call rates all six personas on a question, at about \$0.03 per $1{,}000$ questions in our runs. We center the score $g(p,q)$ over the personas and divide it by its standard deviation on the development questions (Appendix~\ref{app:rater}). With $e_p$ the indicator vector of persona $p$, $\phi(p,q)=[e_p^\top,g(p,q)]^\top\in\mathbb{R}^7$. A user's weights on the indicators express a constant preference for each persona, and the weight on $g$ expresses how closely the user's choices follow the rater's scores.

\textbf{Simulated users.}
Student profiles describe background and preferences in English. We use eight hand-written profiles for development and eight held-out ones for evaluation (Appendix~\ref{app:profile-similarity}). Composite and Quirky are two evaluation rosters whose profiles combine seven traits, including explanation style, answer length, and tone. Composite uses $350$ distinct combinations among $3{,}072$. Quirky draws traits uniformly and adds two of ten personal quirks, such as disliking numbered lists (Appendix~\ref{app:template}).

\textbf{Judge.}
We use \texttt{gpt-6-luna}, with reasoning turned off, to role-play the student and compare two tutor answers. To account for sensitivity to answer order~\citep{wang2024fair} and A/B labels~\citep{zheng2024selectors}, we cover both orders and both label assignments for each profile, question, and persona pair $(p,p')$, sampling four single-letter choices per prompt. This gives $16$ sampled choices across four calls per comparison. We align the four prompts' smoothed log-odds so that positive values favor $p$ over $p'$, then apply the sigmoid to their mean to obtain $P_i(p\succ p'\mid q)$. Appendix~\ref{app:judge-validity} gives the calculation and diagnostics.

\textbf{Agreement with human choices.}
Two authors judged which answer the described student would prefer on \numAgrPairs{} comparisons, each of them on every comparison. They saw the same student profile, question, and two tutor answers as the judge, with the answers in random order and the judge's output hidden. The judge agrees with each author on \numAgrJudgeRone\% and \numAgrJudgeRtwo\% of comparisons, versus \numAgrRaterAgree\% agreement between the authors. Averaging over the two authors, the paired difference is \numAgrDiff{} percentage points (95\% cell-bootstrap interval: \numAgrDiffLo{} to \numAgrDiffHi{}), which includes zero. The judge's confidence predicts how often they choose its favorite answer (Appendix~\ref{app:human}).

\textbf{Second judge.}
We also evaluate a second version of \benchmark{} with TypeSafe Jev as the judge, using the same questions, tutor answers, and student profiles. We fit new evaluation utilities to Jev's choice probabilities. Because Jev also provides the rater scores in the features, we use \texttt{gpt-6-luna} as the rater in this version, so that the judge and the rater remain different models. Comparison methods are retuned on the corresponding development roster (Appendix~\ref{app:second-judge}). On the \numAgrPairs{} author-labeled comparisons, the judges agree with each other on \numAgrJudgesAgree\%, and Jev matches \texttt{gpt-6-luna}'s observed average agreement with the two authors (\numAgrJevAgree\%; Appendix~\ref{app:human}).

\textbf{Evaluation utilities.}
Strong regret~(\ref{eq:regret}) scores both displayed answers against the best one, so it needs a utility for every persona. For each user $i$ and question $q$, we fit one BTL score $\widehat v_p$ per persona to the $15$ pairwise probabilities $P_i(p\succ p'\mid q)$ by ridge-regularized maximum likelihood (Appendix~\ref{app:utilities}), and set
\begin{equation}
\label{eq:exp-utility-scale}
U_i(p,q)=1-\bigl(\max_{p'\in[K]}\widehat v_{p'}-\widehat v_p\bigr)/\kappa,
\end{equation}
with one constant $\kappa=8$ logits for every user and roster, chosen on the development roster; the benchmark score does not depend on $\kappa$ (Appendix~\ref{app:utilities}).

\textbf{Rosters.}
The development roster has $300$ user arrivals drawn from the eight development profiles. The test dataset contains $350$ packs of $24$ questions and their tutor answers, on topics disjoint from development. The held-out-topic roster assigns these packs to users drawn from the same eight profiles. Three further evaluation rosters reuse the packs with new students: the held-out-profile roster assigns each arrival one of the eight new profiles, and the Composite and Quirky rosters give each arrival its own student. Across seeds, we keep the assignment of students to question packs and vary the arrival order and the sampled clicks.

\textbf{Negative control.}
We test whether transfer hurts when users share no preference structure, so that there is nothing to transfer. To build such a roster, we shuffle, for each user independently, the persona labels in the choice probabilities and utilities of the development roster, keeping that shuffle fixed throughout the session and leaving features unchanged. A preference learned for one persona from one user can then correspond to a different persona for another user. The aim is to avoid higher regret than the no-transfer ablation, which learns each user independently.

\subsection{References, Ablations, and Other Methods}
\label{sec:exp-policies}

\textbf{References.}
Table~\ref{tab:main} includes random pairs, the distinct-pair floor, and the per-user linear oracle (Section~\ref{sec:exp-protocol}). Its non-personalized reference is the best fixed pair, chosen in hindsight to minimize regret on the development roster and shown to every user and question. The best fixed pair per user instead chooses a separate pair in hindsight to minimize regret over each user's questions.

\textbf{Ablations and other methods.}
Our \emph{no-transfer ablation} retains only \method{}'s per-user Bayesian logistic learner (Section~\ref{sec:algorithm}), starting each user from the same isotropic prior and learning only from that user's choices. Our \emph{no-recalibration ablation} uses the rater's scores without learning a recalibration from completed sessions. Our \emph{pooled baseline} fits one ridge-regularized logistic model to all completed sessions using $\phi$. It selects the two highest-scoring personas for each question and does not update within a session. Meta-TS~\citep{kveton2021meta} and HierTS~\citep{hong2022hierarchical} learn a prior across bandit tasks. Both model users' preference vectors as Gaussian draws around an unknown population mean, learn this mean from completed sessions, and select pairs by Thompson sampling. Meta-TS samples a prior mean at the start of each session, and HierTS samples it every round from a posterior that also conditions on the current session's choices. We adapt both to pairwise choices by showing the two personas with the highest sampled utilities, and we tune their prior variances like every comparison method (Appendix~\ref{app:results}). Drawing the second persona from a second posterior sample instead, as in double Thompson sampling, moves their excess regret by at most \numTsPairMaxPct\% in either direction on each evaluation roster, a small fraction of their distance to \method{} (Appendix~\ref{app:results}). Appendix~\ref{app:results} also reports AdaTS~\citep{basu2021no}. Our \emph{CoLSTIM} baseline~\citep{bengs2022stochastic} learns each user independently from that user's choices. Before each question it adds random noise to every persona's estimated utility, with more noise for personas whose utility it knows less well, and shows the persona that ranks first after the noise. The noise has the distribution that generates choices in the Bradley--Terry model, so this first pick imitates how a user would choose. The second persona is the one with the best optimistic chance of beating the first, judged by an upper confidence bound on their utility difference. We tune the noise scale together with its other constants. COLDB~\citep{wang2025online} groups users by estimated preferences and pools each group's comparisons, and our variant COLDB-LR assigns a user to a cluster by a likelihood-ratio test. RUCB~\citep{zoghi2014relative} is a non-contextual dueling bandit with one preference matrix pooled over all users and questions. Random pairs select uniformly among distinct persona pairs.

%% file: sections/algorithm.tex
\section{Method}
\label{sec:algorithm}

We propose \method{} (collaborative dueling), which learns from completed sessions to adapt faster to each new user. A \emph{population model} describes users' preference vectors as a mixture of Gaussians fitted to the completed sessions. For a new user, the population model is the prior of a \emph{population learner}, which updates its posterior after every choice and selects pairs with a \emph{lookahead pair rule}: it accepts a pair with higher regret now when the user's choice is expected to lower regret later in the session. An \emph{evidence gate} lets the population learner select pairs only while it predicts completed users' choices better than a \emph{no-transfer learner}, which learns each user alone and selects pairs otherwise. Both learners observe every comparison and, in \benchmark{}, use features $\phi(p,q)$ whose rater score is recalibrated from completed sessions, as described in a paragraph below. Algorithm~\ref{alg:population-transfer} in Appendix~\ref{app:coduel-impl} gives the full procedure.

\textbf{No-transfer learner.}
This learner is Bayesian logistic regression on the current user's choices alone. Each session starts from the prior $\mathcal{N}(0,v_0 I_d)$ with $v_0=B_\theta^2/d$. After each choice, we update a Laplace approximation of the posterior, a Gaussian centered at its mode~\citep[Section~4.4]{bishop2006pattern}, under the likelihood~(\ref{eq:btl}) with $\beta=1$ in \benchmark{} and the generating value in the simulator. The learner shows the two personas with the highest posterior mean utility $\phi(p,q_t)^\top\widehat\theta_t^{\mathrm{nt}}$, which minimizes the current round's expected strong regret under this posterior.

\textbf{Population model.}
The population learner uses the same linear utility $\theta_i^\top\phi(p,q)$ as the no-transfer learner and learns a prior over $\theta_i$ from completed sessions. After $n$ completed sessions, let $\lambda_n^{\mathrm{EM}}$ be the isotropic component's weight fitted by the expectation-maximization (EM)-style procedure below. The prior for a new user is the mixture
\begin{equation}
\label{eq:transfer-prior}
\Pi_n=\lambda_n\,\mathcal{N}(0,v_0 I_d)+(1-\lambda_n)\sum_{k=1}^{M_n}\pi_k\,\mathcal{N}(\mu_k,\Sigma_k),
\qquad \lambda_n=\max\!\left\{\lambda_n^{\mathrm{EM}},\ \frac{\alpha}{\alpha+n}\right\}.
\end{equation}
Its $M_n$ learned components, with weights $\pi_k$, means $\mu_k$, and covariances $\Sigma_k$, describe recurring preference patterns. The isotropic component keeps prior mass for users unlike any seen so far. Its weight $\lambda_n$ is at least $\alpha/(\alpha+n)$, the probability of a new cluster under a Dirichlet process with concentration $\alpha$~\citep{ferguson1973bayesian}. Initially $\Pi_0=\mathcal{N}(0,v_0 I_d)$.

After each session, we refit the mixture to all completed sessions using EM-style updates. We compute each user's responsibility for each component from a Laplace approximation of that user's marginal likelihood under it. We then fit each learned component by ridge-regularized logistic regression on the comparisons of all users, weighted by their responsibilities; the fit's mode and Laplace covariance give $\mu_k$ and $\Sigma_k$. We add a component when a session is better explained by the isotropic component than by any learned one, and remove weakly supported components (Appendix~\ref{app:coduel-impl}).

\textbf{Within-user adaptation.}
The population learner starts each new user from $\Pi_n$, which stays fixed during the session. After each choice, it updates a Laplace posterior within each component from the user's comparisons so far, and reweights the components by how well each explains these choices: each weight becomes proportional to the component's prior weight times its approximate marginal likelihood of the session. Before round $t$, component $k$ has weight $\pi_{k,t}$, mean $m_{k,t}$, and covariance $C_{k,t}$, and this mixture is the learner's \emph{belief} about the user. Component $k$ predicts the utility $s_{k,t}(p;q)=m_{k,t}^\top\phi(p,q)$ of persona $p$ on question $q$, and the probability $\rho_{k,t}(p,p')$ that $p$ is chosen over $p'$ by averaging the likelihood~(\ref{eq:btl}) over its posterior (Appendix~\ref{app:coduel-impl}).

\textbf{Score recalibration.}
In \benchmark{}, the original persona-question features contain the rater's score $g(p,q)$ as their last coordinate (Section~\ref{sec:exp-llm}). A rater can recognize some needs better than others, so its scores need not weigh the six personas as students do. \method{} therefore learns a linear recalibration of these scores from completed sessions. With $g(q)\in\mathbb{R}^K$ the scores of question $q$, it replaces $g(p,q)$ by the $p$-th entry of $A\,g(q)$, where $A=aI_K+A'$. We fit $a$ and $A'$ by logistic regression on all choices of the completed sessions under the likelihood~(\ref{eq:btl}), in which a persona's utility is its recalibrated score plus a persona offset shared by all users and a per-user persona offset. A ridge penalty of $10$ on $A'$ keeps early fits close to a multiple of the rater's scores. We set $A=I_K$ until $20$ sessions are complete and refit it after every $10$ further sessions. After each refit, we rescale the recalibrated coordinate to unit standard deviation over the logged rounds, and both learners rebuild their features, posteriors, and population model from the stored choices. Because $A$ is shared by all users, it carries what completed sessions reveal about the questions, while $\theta_i$ still describes each user. In the simulator, whose features contain no rater score, \method{} runs without this step.

\textbf{Lookahead pair rule.}
Under component $k$, a pair $(p,p')$ on question $q$ has estimated strong regret $r_{k,t}(p,p';q)=2\max_{a\in[K]} s_{k,t}(a;q)-s_{k,t}(p;q)-s_{k,t}(p';q)$. Showing the pair with the lowest belief-averaged regret $\bar r_t(p,p')=\sum_k\pi_{k,t}\,r_{k,t}(p,p';q_t)$ is greedy: it ignores what the user's choice will reveal. When components disagree, a pair whose outcome tells them apart can lower regret for the rest of the session, even if it costs more now. \method{} therefore scores each pair with a one-step lookahead. For each possible choice $y$, it updates the component weights by Bayes' rule to $\pi^{(y)}_{k,t}$, and estimates the regret of greedy pairs over the $h_t=i_tb-t$ remaining rounds of the session. Since future questions are unknown, it averages over a bank $\mathcal{Q}_t$ of the most recent questions across sessions, including $q_t$:
\begin{equation}
\label{eq:transfer-gate}
\begin{aligned}
J_t(p,p')&=\bar r_t(p,p')+h_t\,\mathbb{E}_{y}\!\left[\frac{1}{|\mathcal{Q}_t|}\sum_{q\in\mathcal{Q}_t}\min_{a\ne a'}\sum_k\pi^{(y)}_{k,t}\,r_{k,t}(a,a';q)\right],\\
(p_{t,1},p_{t,2})&\in\arg\min_{p\ne p'}J_t(p,p'),
\end{aligned}
\end{equation}
where the expectation over $y$ uses the belief's predicted probability $\sum_k\pi_{k,t}\rho_{k,t}(p,p')$. Only the component weights change inside the lookahead; the component posteriors stay fixed. This is a knowledge-gradient rule~\citep{frazier2008knowledge,ryzhov2012knowledge} adapted to strong regret. In the last round $h_t=0$, and the rule is greedy.

\textbf{Evidence gate.}
A population fitted to users who share little with the next one can pull that user's estimate in the wrong direction. The gate therefore checks, on completed sessions, whether the population learner predicts choices better than the no-transfer learner. Before each choice is revealed, both learners predict it. For user $i$, let $\ell_i$ be the log-likelihood ratio of the population learner's predictions to the no-transfer learner's, summed over the user's choices. After each session, the gate opens if at least $n_{\min}$ sessions are complete and the mean of $\ell_1,\ldots,\ell_n$ exceeds $z$ standard errors, and closes otherwise. The population learner's advantage can be small when the displayed answers have high but similar utilities, so we use a permissive gate, $z=0.5$ and $n_{\min}=8$, chosen on the development roster. All main comparisons use $\alpha=20$, at most $M_{\max}=16$ learned components, and a question bank of $16$ questions (sensitivity in Appendix~\ref{app:ablations}).

%% file: sections/results.tex
\section{Results}
\label{sec:results}

Throughout, we report the benchmark score of Section~\ref{sec:exp-protocol}, the share of random pairs' excess regret that a policy removes. The tables also give the excess regret $R_T-G_T$.

\subsection{Benchmark results}
\label{sec:res-main}

\input{generated/tab_main}

Table~\ref{tab:main} reports every method on the three evaluation rosters, whose topics and students were never used in design. The held-out-profile roster follows eight new hand-written profiles, and the Composite and Quirky rosters each have $350$ students composed from a template of traits, with two personal quirks added in Quirky. Every policy starts from the first arrival with no preference data. \method{} has the lowest excess regret on all three rosters, with scores of \numUMethodGap{}, \numCMethodGap{}, and \numIMethodGap{}, against \numUBestOtherGap{}, \numCBestOtherGap{}, and \numIBestOtherGap{} for the best comparison method. Rerun on Composite and Quirky rosters redrawn with replacement from the same $350$ students, it scores \numCRsMethodGap{} (\numCRsMethodGapLo{} to \numCRsMethodGapHi{}) and \numIRsMethodGap{} (\numIRsMethodGapLo{} to \numIRsMethodGapHi{}), close to its scores on the actual rosters and on average \numCIRsOverClosest{} points above COLDB, the closest comparison method on both rosters (Appendix~\ref{app:results}, Table~\ref{tab:resampled}). In the second-judge evaluation introduced in Section~\ref{sec:exp-llm}, \method{} again has the lowest excess regret on all three rosters, with \numJUClosestShareAbs{}, \numJCClosestShareAbs{}, and \numJIClosestShareAbs{} less excess regret than the best comparison method on each roster (Appendix~\ref{app:second-judge}).

\textbf{Distance to the oracle.}
The per-user linear oracle, which chooses each user's parameters in hindsight on the features of \method{} (Section~\ref{sec:exp-protocol}), removes \numULinGap\%, \numCLinGap\%, and \numILinGap\% of random pairs' excess regret, so \method{} reaches \numTabMainMethodOfLinLo\% to \numTabMainMethodOfLinHi\% of the oracle's benchmark score. The best fixed pair per user removes \numUPerUserGap\%, \numCPerUserGap\%, and \numIPerUserGap\% (Appendix~\ref{app:results}, Table~\ref{tab:references}), so the oracle's advantage comes from adapting the pair to each question.

\textbf{New topics and new students.} On the held-out-topic roster, which keeps the development student profiles and changes only the questions, \method{} scores \numTMethodGap{}; new students lower its score by \numNewStudentsMethodDropLo{} to \numNewStudentsMethodDropHi{} points (Appendix~\ref{app:results}).

\textbf{Comparison methods.}
Relative to each comparator's excess regret, \method{} reduces excess regret by \numUClosestShareAbs{} against CoLSTIM on the held-out profiles and by \numCClosestShareAbs{} and \numIClosestShareAbs{} against COLDB on Composite and Quirky. These are the strongest comparison methods on the respective rosters. CoLSTIM learns each user separately and scores close to the no-transfer ablation on every roster. On the held-out profiles, it has a higher benchmark score than every comparison method that shares information across users. Meta-TS and HierTS learn a population mean from completed users and both score \numUMetaTsGap{} there, below the no-transfer ablation (\numUNoTransferGap{}), which suggests that their priors can pull users toward a population mean that fits them poorly. Methods that fit one model to all users, the pooled baseline and RUCB, show every student similar pairs and score lowest on the held-out profiles. Appendix~\ref{app:results} gives the paired comparisons (Table~\ref{tab:main-gains}) and the tuning grids, evaluated on development seeds separate from those used for benchmark evaluation (Table~\ref{tab:tuning}).

\textbf{Ablations.}
The learned population matters most: removing it while retaining recalibration raises excess regret by \numAblHNoPopRangeLo\% to \numAblHNoPopRangeHi\% relative to full \method{}. Removing recalibration raises it by \numAblHNoRecalRangeLo\% to \numAblHNoRecalRangeHi\%, a statistically significant increase on each evaluation roster (95\% paired intervals exclude zero). The evidence gate limits harmful transfer in the simulator's null population, where users share no structure: without the gate, excess regret is \numGateSimNullUngatedShareAbs{} higher than no transfer, while \method{} stays within noise of it. On the development roster, which we use for design decisions, lookahead has slightly lower mean excess regret than greedy pairs (\numAblGreedyShareNeg{}), but the difference is not statistically significant. A single Gaussian with learned covariance performs as well as the mixture (\numAblOneGaussShare{}). What matters is that the population describes how users spread around their average: a single tight type without that spread raises excess regret by \numAblOneCompShareAbs{} on the development roster and by \numAblHOneCompRangeLo\% to \numAblHOneCompRangeHi\% on the evaluation rosters. Appendix~\ref{app:ablations} reports every component change on the development roster and on the three evaluation rosters.

\subsection{Learning over the arrivals}
\label{sec:res-phases}

\begin{figure}[t]
\centering
\includegraphics[width=\textwidth]{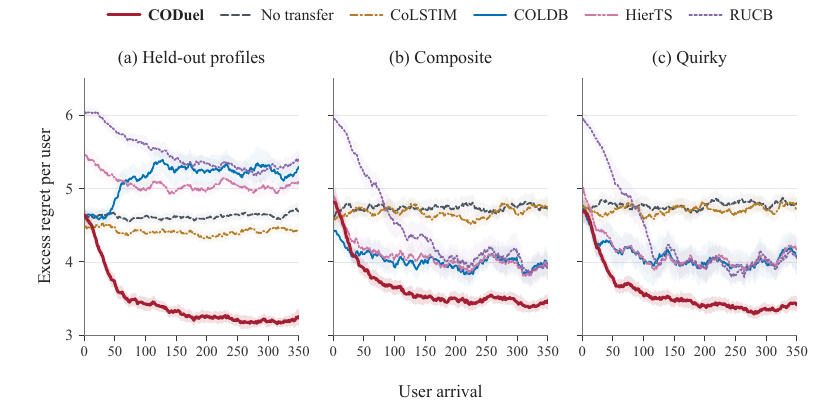}
\caption{Excess regret per user as users arrive on the three evaluation rosters ($b=24$): each user's excess regret averaged over seeds and smoothed over neighboring arrivals, with 95\% bands.}
\label{fig:per-user}
\end{figure}

Figure~\ref{fig:per-user} shows each method's excess regret per user as users arrive. Table~\ref{tab:main} sums over the whole run, including the first users, for whom \method{} has no population to learn from. Over the whole run, \method{} has \numUShareAbs{}, \numCShareAbs{}, and \numIShareAbs{} less excess regret than its no-transfer ablation (last row of Table~\ref{tab:main}; every 95\% interval lies above \numShareAbsMinLo\%, Appendix~\ref{app:uncertainty}). The ablation's excess regret per user stays approximately constant, as expected because it learns each user independently. RUCB also learns across arrivals, but pools all users' choices into one preference matrix without modeling individual preferences. \method{} learns each user alone until the gate turns transfer on, after at least eight users, and then improves as its mixture forms: its excess regret per user is \numEarlyGainLo\% to \numEarlyGainHi\% below the ablation's in the first quarter of the arrivals and \numLateGainLo\% to \numLateGainHi\% below in the last (Figure~\ref{fig:per-user}; Appendix~\ref{app:results}, Table~\ref{tab:phases}).

\subsection{Negative controls and the simulator}
\label{sec:res-maps}

A population model can increase regret when its learned patterns do not apply to a new user. We test \method{} on two negative controls (Appendix~\ref{app:ablations}, Table~\ref{tab:controls}): the simulator's isotropic null population and a scrambled development roster. In the latter, we independently permute each user's persona labels in the choice probabilities and utilities, keeping the permutation fixed throughout the session and leaving features unchanged. This disrupts shared persona preferences while preserving each user's preference patterns under relabeling. On both controls, \method{}'s excess regret stays within one percent of its no-transfer ablation's, with 95\% intervals that include zero. Without its gate, the same policy raises excess regret by \numGateSimNullUngatedShareAbs{} on the simulator's null population and stays within noise on the scrambled roster (Appendix~\ref{app:ablations}, Table~\ref{tab:ablation-gate}). We set the gate's threshold on the development roster, before running either control.

\textbf{Simulator.} Where linear utilities and BTL choices hold exactly, \method{} has lower excess regret than every comparison method in all \numSimAnchorCells{} structured settings. It gains most when users form a few discrete types and least when they form a Gaussian cloud (Appendix~\ref{app:results}, Table~\ref{tab:sim}).

%% file: generated/tab_main.tex
\begin{table}[t]
\centering
\small\setlength{\tabcolsep}{3pt}
\resizebox{\linewidth}{!}{%
\begin{tabular}{lrrrrrr}
\toprule
method & \multicolumn{2}{c}{held-out profiles} & \multicolumn{2}{c}{Composite} & \multicolumn{2}{c}{Quirky} \\
 & excess & score & excess & score & excess & score \\
\midrule
\textit{random pairs} & 2317.3 & 0 & 2276.1 & 0 & 2305.5 & 0 \\
\textit{best fixed pair} & 2197.5 & 5 & 1900.0 & 17 & 1790.3 & 22 \\
\textit{best fixed pair per user (hindsight)} & 546.7 & 76 & 763.2 & 66 & 726.6 & 68 \\
\textit{per-user linear oracle} & 356.8 & 85 & 427.8 & 81 & 411.8 & 82 \\
\textit{distinct-pair floor $G_T$} & 0.0 & 100 & 0.0 & 100 & 0.0 & 100 \\
\midrule
RUCB (pooled, non-contextual) & 1918.5 $\pm$ 5.9 & 17 & 1557.1 $\pm$ 8.1 & 32 & 1528.8 $\pm$ 7.0 & 34 \\
CoLSTIM & 1546.0 $\pm$ 3.6 & 33 & 1636.2 $\pm$ 4.8 & 28 & 1649.3 $\pm$ 4.2 & 28 \\
COLDB & 1801.4 $\pm$ 22.1 & 22 & 1407.2 $\pm$ 11.7 & 38 & 1428.0 $\pm$ 28.2 & 38 \\
HierTS & 1773.4 $\pm$ 4.4 & 23 & 1426.2 $\pm$ 7.5 & 37 & 1438.1 $\pm$ 12.6 & 38 \\
\method{} & \textbf{1197.8 $\pm$ 5.1} & \textbf{48} & \textbf{1276.3 $\pm$ 6.1} & \textbf{44} & \textbf{1250.2 $\pm$ 5.4} & \textbf{46} \\
\midrule
\emph{ablation:} no transfer & 1617.2 $\pm$ 2.9 & 30 & 1658.7 $\pm$ 5.5 & 27 & 1670.9 $\pm$ 4.6 & 28 \\
\bottomrule
\end{tabular}
}
\caption{Benchmark results on the three rosters of new students on new topics ($u=350$ arrivals each, $b=24$, students and topics never used in design). Excess regret $R_T-G_T$ summed over the run (lower is better), mean $\pm$ s.e.\ over seeds (Appendix~\ref{app:uncertainty}), and the benchmark score, the share of random pairs' excess regret that a policy removes (higher is better; Section~\ref{sec:exp-protocol}). Italic rows are evaluation references (Section~\ref{sec:exp-policies}). The distinct-pair floor $G_T$, which scores $100$, follows every question-specific preference, including judge disagreement between display orders, and is not an attainable target; the per-user linear oracle, fitted in hindsight on \method{}'s features, indicates what a learner linear in these features can reach. The no-transfer ablation tests the contribution of learning across users (Section~\ref{sec:res-main}). The floor $G_T$, already subtracted from every entry: 779 (held-out profiles), 699 (Composite), 679 (Quirky).}
\label{tab:main}
\end{table}

%% file: sections/limitations.tex
\section{Limitations and conclusion}
\label{sec:limitations}

\textbf{Evaluation and generalization.}
In \benchmark{}, reference utilities are fitted from the same LLM judge's pairwise probabilities that generate feedback. Human judgments are limited to two authors' choices on \numAgrPairs{} comparisons (Appendix~\ref{app:human}); the judge agrees with them at least as often as they agree with each other. On the three evaluation rosters, Jev and \texttt{gpt-6-luna} favor the same answer in \numSJAgree\% of comparisons excluding ties, and \method{} ranks first under both judges (Appendix~\ref{app:second-judge}). We have not measured human satisfaction or educational outcomes. What students prefer need not be what best supports learning~\citep{bjork2011making,pashler2008learning,puech2025towards}. The benefits of transfer vary across held-out profiles, so average gains do not imply improvements for every profile (Table~\ref{tab:profiles}). Reported intervals do not account for the adaptive development process, and component studies on the development roster are exploratory (Appendix~\ref{app:ablations}).

\textbf{Modeling assumptions and guarantees.}
We assume a stationary population and fixed preference vectors within each session. Linear utilities and BTL feedback approximate the judge's preferences, and Laplace inference and lookahead are approximate (Section~\ref{sec:algorithm}). We provide no regret bound or guarantee against harmful transfer, including under preference drift, misspecified population models, or shifts in the question distribution.

\textbf{Conclusion.}
We introduced \benchmark{}, a benchmark for online LLM personalization from pairwise choices, and \method{}, a baseline that learns a population prior from completed sessions. On students and topics held out from development, \method{} removes \numTabMainMethodGapLo--\numTabMainMethodGapHi\% of random pairs' excess regret, more than the other evaluated methods, and achieves a little over half the performance of a per-user linear oracle. Future work should narrow this gap and evaluate the approach with human users.
\label{sec:conclusion}

%% file: sections/statements.tex
\section*{Ethics Statement}
This work involves no human participants other than the authors. The student preferences in \benchmark{} are simulated: an LLM role-plays students described by short written profiles, and the only human judgments are choices made by two authors to check the judge (Appendix~\ref{app:human}). We collected no personal data. Simulated preferences can differ from those of real students, and an LLM judge can carry biases from its training data, for example by associating a profile's stated background with a teaching style. We therefore present \benchmark{} as a testbed for learning algorithms rather than as evidence of what real students prefer. The system we study adapts the style of answers to the choices a student makes, and the style a student prefers need not be the one that best supports learning~\citep{bjork2011making,pashler2008learning}. A deployed tutor should combine preference feedback with measures of learning and let students see and change the persona serving them. The questions and answers in the dataset were generated by LLMs and may contain errors, so we release them for research on personalization algorithms and not as teaching material.

\section*{Reproducibility Statement}
\ifdefined\codeurl We release the \benchmark{} dataset on \href{\dataurl}{Hugging Face} and, on \href{\codeurl}{GitHub}, the code and commands that rerun every experiment in the paper\else The supplementary material contains the \benchmark{} dataset, the code and commands that rerun every experiment in the paper\fi, the tuning grids and selection rule of the comparison methods, and a script that computes the benchmark scores and confidence intervals from the run results. The dataset stores the questions, the tutor answers, the persona-question features, and the judges' sampled choices and probabilities, including those of the second judge (Appendix~\ref{app:second-judge}), so every experiment runs on CPUs without calling an LLM API. Rebuilding the dataset itself would require the LLMs named in Section~\ref{sec:experiments}. That section describes how the dataset was built, and Appendices~\ref{app:data-quality} and~\ref{app:prompts} give the tutor prompts, the question cues, the student profiles and templates, the judge's probability calculation, and the fit of the evaluation utilities. Section~\ref{sec:algorithm} and Appendix~\ref{app:coduel-impl} specify \method{}, with pseudocode in Algorithm~\ref{alg:population-transfer}. Appendix~\ref{app:simulator} defines the simulated environments, Appendix~\ref{app:results} lists the grids used to tune the comparison methods, and Appendix~\ref{app:uncertainty} describes the confidence intervals. All policies share the same users, questions, and random draws for a given seed, and every setting of \method{} was fixed on the development roster.

\section*{Use of Large Language Models}
LLMs are part of the data and the evaluation of this work. \texttt{gpt-5.6-sol} generated the questions, \texttt{gpt-5.6-terra} generated the tutor answers, \texttt{gpt-6-luna} role-plays the students and provides the pairwise judgments. The rater model TypeSafe Jev scores the personas' answer styles for the features (Section~\ref{sec:exp-llm}) and judges the evaluation rosters a second time, with \texttt{gpt-6-luna} as the rater in that evaluation (Appendix~\ref{app:second-judge}). \texttt{gpt-5.6-sol} and \texttt{claude-opus-5} serve as comparison judges (Appendix~\ref{app:data-quality}). An LLM drafted the tutor persona prompts, the student profiles, the trait templates, and the question cues, and the authors edited them and selected the final versions. We used Claude Code, Cursor, and ChatGPT and Codex to implement the benchmark, the comparison methods, and \method{}, and to run and analyze experiments. Coding agents also proposed and tested variants of \method{} on the development roster and the tuning seeds, under an evaluation protocol that the authors fixed in advance, and the authors selected the reported configuration. LLMs drafted and edited parts of the text, and the authors revised and approved every sentence. LLMs also helped search for related work and check the cited references. The authors defined the problem and the evaluation protocol, reviewed all code and text, and checked every reported number with scripts that recompute it from the stored results. The authors take full responsibility for the content of the paper.

%% file: sections/appendix_simulator.tex
\section{Simulator constructions}
\label{app:simulator}

The simulator uses linear utilities and logistic pairwise feedback, as in the contextual preference-bandit models of \citet{saha2021optimal} and \citet{bengs2022stochastic}. Our $\phi(p,q_t)$ corresponds to the contextual feature vector for arm $p$ at round $t$. The feature construction and user populations below are specific to our simulator. Implementation details and reported configurations are available in the accompanying code (\texttt{persona\_bench/world.py} and \texttt{results/worlds.py}).

We use $d=8$, $K=6$, and $B_\theta=1$. A question is a unit vector $z_q\in\mathbb{R}^d$. Each persona has a matrix $B_p\in\mathbb{R}^{d\times d}$ with independent $\mathcal{N}(0,1/d)$ entries, drawn once per world and then fixed. The matrix $B_p$ transforms the shared question vector into features specific to persona $p$, allowing persona rankings to vary across questions. We normalize the resulting vector to obtain the feature map
\begin{equation}
\label{eq:sim-features}
\phi(p,q)=\frac{B_pz_q}{\|B_pz_q\|_2}.
\end{equation}
Utilities and choices follow (\ref{eq:linear}) and (\ref{eq:btl}); policies receive the true $B_\theta$, $d$, and $\beta$. Default questions are i.i.d.\ directions on the sphere.

Write $\Pi(v)=v/\max\{1,\|v\|_2/B_\theta\}$ for projection onto the norm ball. Users are sampled independently. The learner is not told which population is active.

\textbf{Finite types.}
For $m\in\{2,4,8\}$, user $i$ draws a type $h_i$ uniformly from $[m]$ and has $\theta_i=\theta^{h_i}$. The type vectors have norm $B_\theta$ and equal pairwise angles, either $60^\circ$ or $90^\circ$, and are constructed from a random orthonormal frame.

\textbf{Gaussian cloud.}
We draw $v_i\sim\mathcal{N}(\mu_{\mathrm{pop}},s^2C_d)$ and set $\theta_i=\Pi(v_i)$, with $s=0.3$. The center has norm $B_\theta/2$ and a uniformly random direction. The random positive-definite matrix $C_d$ is normalized to have trace $d$.

\textbf{Low-rank cloud.}
We use this population to test transfer when variation across users is concentrated in a shared low-dimensional subspace. The latent directions represent correlated changes in users' preference weights. Before projection, the covariance has the standard low-rank plus isotropic noise form of probabilistic principal component analysis~\citep{tipping1999probabilistic}. We use $r=2$ and $s=1$, drawing
\begin{equation}
\label{eq:sim-lowrank}
v_i\sim\mathcal{N}\!\left(\mu_{\mathrm{pop}},s^2\left(VC_rV^\top+0.05^2 I_d\right)\right),
\qquad \theta_i=\Pi(v_i).
\end{equation}
Here $V\in\mathbb{R}^{d\times r}$ has orthonormal columns spanning a uniformly random subspace, $C_r$ is positive definite with trace $r$, and $\mu_{\mathrm{pop}}$ has norm $B_\theta/2$ and lies in that subspace. The isotropic component has standard deviation $0.05s$ and allows small deviations outside the shared subspace, so the full covariance is not rank deficient.

\textbf{Null population.}
Users are drawn from $\mathcal{N}(0,(B_\theta^2/d)I_d)$ and projected with $\Pi$. This isotropic negative control contains no additional population structure.

\textbf{World screening.}
On $256$ independently sampled diagnostic questions we measure whether users disagree on their favorite persona. We regenerate a world if fewer than $5\%$ of questions elicit any disagreement. For finite types the check uses every type; for either cloud or the null population it uses $64$ independent population draws. These diagnostic observations are not supplied to the policies.

\textbf{Reported configurations.}
Each configuration uses $u=300$ users. Table~\ref{tab:sim} reports the $18$ anchor configurations at $b=20$ and $\beta\in\{4,12\}$, including the two null configurations. These experiments use i.i.d.\ questions. Appendix~\ref{app:results} summarizes a broader grid over session length $b$ and click consistency $\beta$.

%% file: sections/appendix_uncertainty.tex
\section{Uncertainty estimation}
\label{app:uncertainty}

For simulator paired gains, we report $95\%$ Student's $t$ confidence intervals across seeds. For the main tutoring comparisons, we report confidence intervals from a crossed bootstrap that independently resamples seeds and users. On rosters with recurring profiles (the held-out-profile roster, the held-out-topic roster and the development roster), users are resampled within profiles; on the Composite and Quirky rosters every user has a profile of their own, and users are resampled without strata. Bootstrap intervals use the $2.5$th and $97.5$th percentiles, with $2{,}000$ replicates. This calculation reuses recorded per-user regret and does not rerun the adaptive policies. The intervals are therefore conditional on the evaluated students: a user's regret depends on the population learned from earlier users, and resampling users after the fact does not simulate a different population. On the held-out-profile roster users are resampled within its eight profiles and the profiles are not resampled; Table~\ref{tab:profiles} reports results per profile, and Appendix~\ref{app:results} reports a range that also resamples profiles after the fact. Standard errors in the tables are over seeds, with the roster fixed. Table~\ref{tab:main} uses \numSeedsU{} seeds on the held-out-profile roster and \numSeedsC{} on each of Composite and Quirky. Table~\ref{tab:resampled} reruns the policies on a redrawn roster for every seed, which covers the choice of students as well; on the held-out-profile roster, drawing eight profiles with replacement leaves fewer distinct profiles per roster, so that table describes the spread of such rosters rather than an unbiased estimate of the actual roster's score. Exploratory studies (Appendix~\ref{app:ablations}) use standard errors or paired seed-based $t$ intervals, as their captions specify.

%% file: sections/appendix_dataset.tex
\section{Dataset characteristics and quality}
\label{app:data-quality}

These diagnostics describe the development roster, the held-out-topic roster, and the three rosters of new students: the held-out-profile roster, on which every arrival comes from the eight new hand-written profiles, and the Composite and Quirky rosters, on which every arrival has a profile of its own. They characterize the simulated preferences; Appendix~\ref{app:human} compares the judge's choices with human choices.

\subsection{Roster statistics}
\label{app:development-diagnostics}

\input{generated/tab_dataset}

We center each question's six Bradley--Terry scores and decompose them into a persona effect shared by all users, the deviation of each profile's mean from it, the deviation of each session's mean from its profile mean, and the deviation of each question's scores from its session mean (Table~\ref{tab:dataset-quality}). On the development roster, profile--persona differences account for \numDataDVarType\% of the centered score variance and question deviations within sessions for \numDataDVarQuestion\%. Part of this within-session variation reflects the judge's disagreement between display orders (Appendix~\ref{app:judge-validity}). The new profiles on the held-out-profile roster differ more from one another than the development profiles do (\numDataUVarType\% against \numDataDVarType\%). On the Composite and Quirky rosters each profile has a single session, so profile and session deviations cannot be separated; together they account for \numDataCVarType\% and \numDataIVarType\%. Preference concentration is the KL divergence of each stored Bernoulli choice distribution from an unbiased choice, $P\log(2P)+(1-P)\log\bigl(2(1-P)\bigr)$ for a pair with preference probability $P$, averaged over questions and persona pairs in nats. Judgment/utility disagreement is the fraction of comparisons in which the stored preference probability favors the persona with lower fitted utility; it measures how far the judge's pairwise probabilities are from their Bradley--Terry approximation.

\subsection{Judge diagnostics}
\label{app:judge-validity}

\textbf{Choice probabilities.}
We derive probabilities from sampled choices rather than token probabilities~\citep{wang2024answerc}, which a chat model need not keep calibrated after post-training~\citep{openai2023gpt4}. If $n_A$ of a prompt's four choices are \texttt{A}, its smoothed log-odds are $\log\bigl((n_A+\tfrac12)/(4-n_A+\tfrac12)\bigr)$. We reverse the sign when \texttt{A} denotes $p'$, average over the four prompts, and apply the sigmoid to obtain $P_i(p\succ p'\mid q)$.

The benchmark's choice probabilities are measurements made by the judge. Table~\ref{tab:judge-validity} characterizes them on the \numJVPrompts{} stored prompts of all $24$ questions of every arrival.

\input{generated/tab_judge_validity}

\textbf{Answer format and sampling.}
All sampled choices were valid labels (\texttt{A} or \texttt{B}). The judge's four choices for a prompt agree on \numJVPromptsAtCapLo\% to \numJVPromptsAtCapHi\% of prompts, whose log-odds then sit at the cap of $\log 9$, and \numJVPairsSaturatedLo\% to \numJVPairsSaturatedHi\% of persona pairs have all $16$ choices on the same side, so the strongest preferences are compressed to a probability of about $0.9$. Splitting each prompt's four choices into two halves gives scores that correlate closely (reliability of the full average \numJVRelPicksLo{} to \numJVRelPicksHi{}): the sampling of choices adds little noise.

\textbf{Agreement of the four prompts.}
Swapping which answer is labeled A changes the judgment little: the log-odds of the two label assignments correlate \numJVRLabelHalvesLo{} to \numJVRLabelHalvesHi{}, although the judge leans \numJVLeanLabelIALo{} to \numJVLeanLabelIAHi{} toward the answer labeled A. The display order matters more. The judge leans \numJVLeanFirstShownLo{} to \numJVLeanFirstShownHi{} toward the answer shown first, the log-odds of the two orders correlate \numJVROrderHalvesLo{} to \numJVROrderHalvesHi{}, and all four prompts agree in sign on \numJVAllFourSameSignLo\% to \numJVAllFourSameSignHi\% of persona pairs. Averaging over the two orders and labels removes the constant leans but not the disagreement specific to each comparison. With scores fitted separately on the two orders, the reliability of the four-prompt average is \numJVRelStudentLo{} to \numJVRelStudentHi{} for a student's profile over a session and \numJVRelQuestionDevLo{} to \numJVRelQuestionDevHi{} for the deviations of each question from its session. Question-specific deviations are less consistent across answer orders than session averages.

\textbf{Length.}
Brevity is part of the concise persona's style. The answer instruction asks it for $40$ to $90$ words and the other personas for $100$ to $180$, and lets every tutor leave that range when the question calls for it (Appendix~\ref{app:tutor-prompts}); on the held-out questions its answers have a median of \numJVWordsConcise{} words, against \numJVWordsOtherLo{} to \numJVWordsOtherHi{} for the other personas. LLM judges can favor longer answers~\citep{dubois2024length}. Holding the persona pair and the student profile fixed, our judge instead slightly prefers the shorter answer (\numJVLenEffectSdLo{} to \numJVLenEffectSdHi{} logits per standard deviation of the log length ratio). Across all comparisons, which mix length with the personas' styles, the longer answer is preferred in \numJVLongerWinsLo\% to \numJVLongerWinsHi\% of them.

\textbf{Consequences for the benchmark.}
All policies face the same fixed table of choice probabilities, including its display-order effects and capped log-odds. Some question-specific variation may be difficult to predict from our features $\phi(p,q)$, so excess regret can reflect limitations of the representation as well as learning.

\subsection{Rater model}
\label{app:rater}

We send the rater one query per question. The query contains the question text and a short description of each persona's answer style, written from the persona's system prompt (Appendix~\ref{app:rater-prompt}). It contains no student profile, tutor answer, or choice. The query asks which style of answer would serve the student's request worst, and the rater returns a probability $\pi(p\mid q)$ for each of the six styles, rounded to two decimals. We set $g(p,q)=-\log\max\{\pi(p\mid q),10^{-3}\}$, so a style the rater considers unlikely to be the worst receives a high score, and the floor $10^{-3}$ handles probabilities reported as zero. We then center $g$ over the six personas of each question and divide it by its standard deviation over all persona-question pairs of the development questions. The test questions use the same constant.

We chose this query on the development roster among three variants: the one above, a positively phrased one asking which style would serve the request best, and one asking, for each persona, whether the student wants what that persona offers. The worst-style query best captured how each user's preferences change from one question to the next.

\subsection{Evaluation utilities}
\label{app:utilities}

We chose $\kappa$ above the largest fitted score range on the development roster. A common scale keeps differences in preference strength: a question on which the student barely distinguishes the personas contributes less regret than one on which the favorite is far ahead. The benchmark score does not depend on $\kappa$. We score regret on $U$ and draw clicks from $P$, so fitting utilities does not change the feedback. The learners use $\beta=1$ and a prior scale $B_\theta=1.84$ fitted on the development roster (Appendix~\ref{app:coduel-impl}).

We fit the BTL scores of Section~\ref{sec:exp-llm} by
\begin{equation}
\label{eq:exp-btl-fit}
\begin{aligned}
\widehat v=\arg\min_{v\in\mathbb{R}^K}\Bigg\{
&-\sum_{1\le p<p'\le K}\Big[P_{pp'}\log\sigma(v_p-v_{p'})\\
&\quad +(1-P_{pp'})\log\sigma(v_{p'}-v_p)\Big]
+\frac{0.05}{2}\|v\|_2^2\Bigg\}.
\end{aligned}
\end{equation}
All $15$ unordered pairs receive equal weight. The ridge term keeps scores finite when probabilities are near $0$ or $1$.

\subsection{Agreement with human choices}
\label{app:human}

\textbf{Labeled comparisons.}
We drew \numAgrSamplePairs{} comparisons: \numAgrSampleUnits{} (student, question) cells, one question per arrival, spread over the five rosters and, within the development, held-out-topic and held-out-profile rosters, balanced over profiles; in each cell $4$ of the $15$ persona pairs, chosen so that the sample covers the judge's range of confidence evenly. Both authors labeled every comparison of \numAgrCells{} cells, \numAgrPairs{} comparisons in total. The interface showed the student profile, the question, and the two answers, exactly as in the judge's prompt, in an independently randomized display order for each author, and asked which answer this student would rather receive; a choice was required. It never showed the judge's output, the persona names, or the other author's choice. The authors made \numAgrLabels{} choices, \numAgrPairs{} each. Throughout, a comparison counts once, with the two authors' choices averaged, and we report 95\% intervals from a bootstrap over (student, question) cells, since comparisons within a cell share a student and a question.

\input{generated/tab_agreement}

\textbf{The judge agrees with each author at least as often as the authors agree with each other.}
The fair reference for the judge is a second human who sees the same comparison. The two authors agree with each other on \numAgrRaterAgree\% of the comparisons (Cohen's $\kappa=\numAgrKappaRaters$; Table~\ref{tab:agreement}a). The judge agrees with them on \numAgrJudgeRone\% and \numAgrJudgeRtwo\% ($\kappa=\numAgrKappaJudgeRone$ and \numAgrKappaJudgeRtwo), \numAgrDiff{} points more often than the authors agree with each other (\numAgrDiffLo{} to \numAgrDiffHi). Where the two authors agree, the judge picks their answer in \numAgrBothAgreeJudge{} of \numAgrBothAgreeN{} comparisons. The judge's log-odds also rank the authors' choices well, with an AUC of \numAgrAuc{} (\numAgrAucLo{} to \numAgrAucHi). Disagreement with the judge is thus of the same size as disagreement between two careful readers of the same profile: which of two good answers a described student prefers is partly a matter of judgment. The second judge of Appendix~\ref{app:second-judge}, TypeSafe Jev, agrees with the authors just as often (\numAgrJevAgree\%, against \numAgrJudgeAgree\% for \texttt{gpt-6-luna}), and the two judges agree with each other on \numAgrJudgesAgree\% of the comparisons, more often than the two authors do.

\textbf{The judge's confidence tracks how clear-cut a comparison is for people.}
A useful judge should be more often right when it is more confident, and its uncertain comparisons should be those on which people also hesitate. Table~\ref{tab:agreement}b groups the comparisons by the probability the judge gives its favorite answer. When that probability is \numAgrBinTopP{} or more, which covers \numAgrBinTopShare\% of the comparisons, the authors choose the judge's favorite in \numAgrBinTopHit\% of cases, close to the stated probability. Between $0.7$ and $0.9$ the judge is over-confident: the authors choose its favorite in \numAgrBinMidHit\% of comparisons, against a stated \numAgrBinMidP{}. Averaged over the bands, the stated probability and the authors' choices differ by \numAgrEce{} points. The authors' agreement with each other follows the judge's confidence closely, from \numAgrBinLowRater\% on the comparisons the judge finds hardest to \numAgrBinCapRater\% at its maximum probability. These rates match what two people would reach if each chose the judge's favorite with the judge's probability (\numAgrBinLowExp\% and \numAgrBinCapExp\%). The judge's uncertainty therefore marks comparisons that are ambiguous for people too, and the clicks it generates reproduce how often two people would disagree.

\textbf{Larger judges agree with the authors about as often.}
Table~\ref{tab:agreement}c scores other judges on the same comparisons. Two larger chat models, \texttt{gpt-5.6-sol} and \texttt{claude-opus-5} (thinking off), are run exactly like the dataset's judge: the same four prompts with four sampled choices each. Their agreement with the authors differs from the dataset judge's by \numAgrDiffSol{} and \numAgrDiffOpus{} points, with intervals that include zero (\numAgrDiffSolLo{} to \numAgrDiffSolHi{} and \numAgrDiffOpusLo{} to \numAgrDiffOpusHi). Where the dataset's judge is confident, \texttt{gpt-5.6-sol} favors the same answer in \numAgrSameFavHighSol\% of comparisons and \texttt{claude-opus-5} in \numAgrSameFavHighOpus\%.

\textbf{Limits.}
The labels come from two authors, not from students of the described kinds. One of the two authors wrote the student profiles.

\subsection{A second judge}
\label{app:second-judge}

The judge's probabilities generate the clicks, and the evaluation utilities are fitted to the same probabilities. To test whether the benchmark depends on this judge, we judged every comparison of the five rosters again with TypeSafe Jev, the rater of Section~\ref{sec:exp-llm}. As for the dataset's judge, each comparison is posed in both display orders and with both assignments of the labels A and B, one call each. Jev returns the probability of each answer, so we directly average the four calls' log-odds. The two judges mostly agree on what each student prefers (Table~\ref{tab:second-judge}). Among the \numSJAgreeComparisons{} comparisons of two personas' answers on the three evaluation rosters, Jev favors the same answer as \texttt{gpt-6-luna} in \numSJAgree\% of those where neither judge is indifferent, and in \numSJAgreeTop\% of those where \texttt{gpt-6-luna} gives its maximum probability of $0.9$ (\numSJAgreeTopShare\% of comparisons). The two authors agree with each other less often, on \numAgrRaterAgree\% of the comparisons they both labeled (Appendix~\ref{app:human}). A click sampled from Jev matches one sampled from \texttt{gpt-6-luna} with probability \numSJClickCross, close to the \numSJClickSelf{} of two clicks from \texttt{gpt-6-luna} and the \numSJClickSelfJ{} of two clicks from Jev. Summed over each student's $24$ questions, the utilities of the two judges give the same most-preferred persona for \numSJTopPersona\% of the \numSJStudents{} students (\numSJTopPersonaChance\% by chance). Choosing each student's pair from \texttt{gpt-6-luna}'s utilities removes \numSJTransferU\%, \numSJTransferC\%, and \numSJTransferI\% of random pairs' excess regret under Jev's utilities on the three rosters, against \numSJTransferOwnU\%, \numSJTransferOwnC\%, and \numSJTransferOwnI\% for the pair chosen from Jev's own.

We then ran every method on the five rosters judged by Jev. Because Jev is also the rater, every method reads the scores of a second rater, \texttt{gpt-6-luna}, so that rater and judge differ. The comparison methods are re-tuned on the development roster judged by Jev, with the grids of Table~\ref{tab:tuning}. \method{}'s prior scale follows the same rule as for the dataset's judge ($B_\theta=2.70$, Appendix~\ref{app:ablations}), and every other setting is unchanged. As with the original judge (Table~\ref{tab:main}), \method{} has the lowest excess regret on all three evaluation rosters (Table~\ref{tab:sj-main}), with scores of \numJUMethodGap{}, \numJCMethodGap{}, and \numJIMethodGap{}, against \numJUBestOtherGap{}, \numJCBestOtherGap{}, and \numJIBestOtherGap{} for the best comparison method, and \numJUClosestShareAbs{}, \numJCClosestShareAbs{}, and \numJIClosestShareAbs{} less excess regret than the closest. The per-user linear oracle scores \numJLinRange{} under Jev, so \method{} achieves \numJMethodOfLinLo\% to \numJMethodOfLinHi\% of its performance. Table~\ref{tab:sj-main-gains} gives paired confidence intervals for gains over the comparison methods. \method{} also ranks first on the held-out-topic and development rosters, with scores of \numJTMethodGap{} and \numJDMethodGap{}, respectively (Table~\ref{tab:sj-topic}). The development results are descriptive.

\input{generated/tab_second_judge}
\input{generated/tab_sj_main}
\input{generated/tab_sj_main_gains}
\input{generated/tab_sj_topic}

%% file: generated/tab_dataset.tex
\begin{table}[t]
\centering
\small
\resizebox{\linewidth}{!}{%
\begin{tabular}{lrrrrr}
\toprule
statistic & development & held-out topics & held-out profiles & Composite & Quirky \\
\midrule
arrivals & 300 & 350 & 350 & 350 & 350 \\
distinct profiles & 8 & 8 & 8 & 350 & 350 \\
\midrule
\multicolumn{6}{l}{\emph{share of centered score variance (\%)}} \\
\quad shared persona effects & 18.0 & 20.5 & 7.1 & 13.1 & 13.0 \\
\quad profile--persona differences & 31.7 & 28.2 & 50.3 & 30.4 & 34.5 \\
\quad sessions within profiles & 2.0 & 2.1 & 1.9 & 0.0 & 0.0 \\
\quad question deviations within sessions & 48.3 & 49.1 & 40.7 & 56.5 & 52.5 \\
\midrule
mean preference concentration (nats) & 0.231 & 0.231 & 0.226 & 0.225 & 0.228 \\
judgment/utility disagreement (\%) & 2.8 & 2.8 & 2.4 & 3.4 & 3.2 \\
\bottomrule
\end{tabular}
}
\caption{Dataset statistics by roster ($24$ questions per arrival). Variance shares decompose question-centered Bradley--Terry scores into a shared persona effect, profile deviations, session deviations within a profile and question deviations within a session; the question deviations include the judge's disagreement between display orders (Appendix~\ref{app:judge-validity}). On the Composite and Quirky rosters every profile has one arrival, so profile and session deviations coincide. Preference concentration is the KL divergence from an unbiased choice, measured in nats (units based on natural logarithms). Concentration and disagreement average over all $15$ unordered persona pairs per question.}
\label{tab:dataset-quality}
\end{table}

%% file: generated/tab_judge_validity.tex
\begin{table}[t]
\centering
\small\setlength{\tabcolsep}{4pt}
\resizebox{\linewidth}{!}{%
\begin{tabular}{lrrrrr}
\toprule
diagnostic & development & held-out topics & held-out profiles & Composite & Quirky \\
\midrule
sampled choices that are a valid label (\%) & 100.0 & 100.0 & 100.0 & 100.0 & 100.0 \\
prompts whose $4$ choices agree (\%) & 77 & 78 & 79 & 73 & 74 \\
persona pairs with all $16$ choices on one side (\%) & 40 & 41 & 41 & 38 & 39 \\
reliability: split of each prompt's choices & 0.99 & 0.99 & 0.99 & 0.99 & 0.99 \\
lean toward the first-shown answer (logits) & +0.38 & +0.38 & +0.48 & +0.24 & +0.24 \\
lean toward the label A (logits) & +0.22 & +0.21 & +0.13 & +0.22 & +0.23 \\
pairs on which all four prompts agree in sign (\%) & 54 & 55 & 54 & 54 & 54 \\
correlation, the two label assignments & 0.89 & 0.89 & 0.90 & 0.89 & 0.89 \\
correlation, the two display orders & 0.72 & 0.70 & 0.65 & 0.71 & 0.70 \\
reliability: question deviations within a session & 0.86 & 0.86 & 0.82 & 0.87 & 0.86 \\
reliability: student profiles & 0.96 & 0.95 & 0.94 & 0.96 & 0.96 \\
length: logit change per s.d.\ of the length ratio & -0.101 & -0.090 & -0.161 & -0.147 & -0.147 \\
length: longer answer preferred (\%) & 48 & 50 & 44 & 52 & 50 \\
\bottomrule
\end{tabular}
}
\caption{Diagnostics of the judge (\texttt{gpt-6-luna}, $4$ sampled choices per prompt) on the stored judgments of all $24$ questions of every arrival (four prompts per persona pair and question). A prompt's log-odds is capped at $\log 9$ when its four choices agree. Correlations are between the logits averaged over one half of the prompts and over the other half, across persona pairs. Reliabilities are Spearman--Brown estimates for the four-prompt average, from Bradley--Terry scores fitted separately on the two display orders: of each question's scores minus the session's mean scores, and of each session's mean scores after removing the persona means. Length: log ratio of the two answers' word counts against the four-prompt logit, both centered within student profile and persona pair.}
\label{tab:judge-validity}
\end{table}

%% file: generated/tab_agreement.tex
\begin{table}[t]
\centering
\small
\resizebox{\linewidth}{!}{%
\begin{tabular}{lrrrr}
\toprule
\multicolumn{5}{l}{\emph{(a) Pairs of raters}}\\
Pair & \multicolumn{3}{r}{Agreement (\%)} & Cohen's $\kappa$\\
\midrule
Author 1 and author 2 & \multicolumn{3}{r}{69.6 [61.8, 77.0]} & 0.39\\
\texttt{gpt-6-luna} and author 1 & \multicolumn{3}{r}{74.0} & 0.48\\
\texttt{gpt-6-luna} and author 2 & \multicolumn{3}{r}{73.0} & 0.46\\
TypeSafe Jev and author 1 & \multicolumn{3}{r}{71.1} & 0.42\\
TypeSafe Jev and author 2 & \multicolumn{3}{r}{76.0} & 0.52\\
\texttt{gpt-6-luna} and TypeSafe Jev & \multicolumn{3}{r}{81.4 [76.0, 86.8]} & 0.63\\
\texttt{gpt-6-luna}, where the authors agree (142) & \multicolumn{3}{r}{83.8} & \\
AUC of the log-odds of \texttt{gpt-6-luna} & \multicolumn{3}{r}{0.79 [0.73, 0.84]} & \\
\midrule
\multicolumn{5}{l}{\emph{(b) By the confidence of \texttt{gpt-6-luna}}}\\
Probability of its favorite & Comparisons & Mean & Authors choose it (\%) & Authors agree (\%), expected\\
\midrule
$[0.5, 0.7)$ & 38 & 0.59 & 61.8 [47.6, 74.5] & 55, 52\\
$[0.7, 0.9)$ & 78 & 0.80 & 66.0 [58.7, 73.6] & 63, 68\\
$0.9$ (maximum) & 88 & 0.90 & 85.2 [77.8, 92.0] & 82, 82\\
\midrule
\multicolumn{5}{l}{\emph{(c) Other judges}}\\
Judge & Comparisons & \multicolumn{2}{r}{Agreement (\%)} & Same favorite (\%)\\
\midrule
\texttt{gpt-6-luna} (this dataset) & 204 & \multicolumn{2}{r}{73.5 [68.4, 78.4]} & --\\
\texttt{gpt-5.6-sol} & 204 & \multicolumn{2}{r}{74.5 [68.6, 79.9]} & 84.8\\
\texttt{claude-opus-5} & 204 & \multicolumn{2}{r}{74.0 [68.6, 79.2]} & 76.0\\
\bottomrule
\end{tabular}}
\caption{Agreement with the authors' choices on the 204 comparisons, each labeled by both authors. Brackets: 95\% intervals of a bootstrap over the 51 labeled (student, question) cells. (a) Agreement between each pair of raters, the two judges included; the judge's favorite is its more probable answer. (b) By the confidence of \texttt{gpt-6-luna}, the probability it gives its favorite (at most $0.9$): how often the authors choose that favorite, how often they agree with each other, and the agreement two raters would reach if each chose the favorite with the judge's probability. (c) Other judges run with the same four prompts; same favorite: share of comparisons on which the judge favors the same answer as \texttt{gpt-6-luna}.}
\label{tab:agreement}
\end{table}

%% file: generated/tab_second_judge.tex
\begin{table}[t]
\centering
\small
\resizebox{\textwidth}{!}{%
\begin{tabular}{lrrr}
\toprule
statistic & held-out profiles & Composite & Quirky \\
\midrule
same favored answer, comparisons (\%) & 84.3 & 78.5 & 79.4 \\
\quad where \texttt{gpt-6-luna} gives $0.9$ (\%) & 96.1 & 92.5 & 92.8 \\
click agreement, Jev with \texttt{gpt-6-luna} & 0.68 & 0.64 & 0.65 \\
\quad two \texttt{gpt-6-luna} clicks & 0.70 & 0.70 & 0.70 \\
\quad two Jev clicks & 0.75 & 0.70 & 0.71 \\
same most-preferred persona, students (\%) & 88 & 72 & 77 \\
score of each student's best pair under \texttt{gpt-6-luna}, paid under Jev & 85 & 73 & 76 \\
\quad its best pair under Jev & 87 & 83 & 83 \\
\bottomrule
\end{tabular}}
\caption{Agreement of the second judge (TypeSafe Jev) with the dataset's judge (\texttt{gpt-6-luna}) on the same comparisons of the three evaluation rosters (\numSJAgreeComparisons{} comparisons of two personas' answers, \numSJStudents{} students). Favored answer: comparisons where neither judge gives $0.5$. Click agreement: probability that a click from each judge picks the same answer. Most-preferred persona: utilities summed over the student's questions. Last rows: benchmark score under Jev's utilities of each student's best fixed pair, chosen from either judge's utilities.}
\label{tab:second-judge}
\end{table}

%% file: generated/tab_sj_main.tex
\begin{table}[t]
\centering
\small\setlength{\tabcolsep}{3pt}
\resizebox{\linewidth}{!}{%
\begin{tabular}{lrrrrrr}
\toprule
method & \multicolumn{2}{c}{held-out profiles} & \multicolumn{2}{c}{Composite} & \multicolumn{2}{c}{Quirky} \\
 & excess & score & excess & score & excess & score \\
\midrule
\textit{random pairs} & 3008.2 & 0 & 2529.8 & 0 & 2551.8 & 0 \\
\textit{best fixed pair} & 2510.0 & 17 & 1329.1 & 47 & 1809.3 & 29 \\
\textit{per-user linear oracle} & 293.4 & 90 & 264.5 & 90 & 259.7 & 90 \\
\textit{distinct-pair floor $G_T$} & 0.0 & 100 & 0.0 & 100 & 0.0 & 100 \\
\midrule
RUCB (pooled, non-contextual) & 2560.3 $\pm$ 8.0 & 15 & 1545.7 $\pm$ 8.0 & 39 & 1866.6 $\pm$ 7.7 & 27 \\
Pooled & 2569.5 $\pm$ 34.7 & 15 & 1772.5 $\pm$ 84.6 & 30 & 2002.9 $\pm$ 57.4 & 22 \\
CoLSTIM & 1686.5 $\pm$ 3.0 & 44 & 1536.4 $\pm$ 4.3 & 39 & 1525.4 $\pm$ 4.8 & 40 \\
COLDB & 1789.4 $\pm$ 15.4 & 41 & 1451.7 $\pm$ 15.2 & 43 & 1524.1 $\pm$ 14.5 & 40 \\
COLDB-LR & 1666.8 $\pm$ 17.2 & 45 & 1528.8 $\pm$ 17.6 & 40 & 1590.0 $\pm$ 17.7 & 38 \\
Meta-TS & 2477.4 $\pm$ 18.0 & 18 & 1477.8 $\pm$ 30.5 & 42 & 1874.5 $\pm$ 19.9 & 27 \\
HierTS & 2415.3 $\pm$ 7.7 & 20 & 1417.5 $\pm$ 7.1 & 44 & 1821.2 $\pm$ 4.6 & 29 \\
\method{} & \textbf{1243.6 $\pm$ 6.3} & \textbf{59} & \textbf{1128.3 $\pm$ 5.9} & \textbf{55} & \textbf{1191.2 $\pm$ 6.2} & \textbf{53} \\
\midrule
\emph{ablation:} no transfer & 1734.4 $\pm$ 4.5 & 42 & 1571.6 $\pm$ 6.3 & 38 & 1543.4 $\pm$ 4.6 & 40 \\
\bottomrule
\end{tabular}
}
\caption{Second judge (TypeSafe Jev): benchmark results on the three rosters of new students on new topics, same layout as Table~\ref{tab:main}, with clicks and evaluation utilities from Jev and \method{} reading the scores of a second rater, \texttt{gpt-6-luna}. Every comparison method is tuned on the development roster judged by Jev (\numJSeedsU{} seeds on the held-out profiles, \numJSeedsC{} on Composite and Quirky). The floor $G_T$, already subtracted from every entry: 977 (held-out profiles), 887 (Composite), 873 (Quirky).}
\label{tab:sj-main}
\end{table}

%% file: generated/tab_sj_main_gains.tex
\begin{table}[t]
\centering
\small\setlength{\tabcolsep}{4pt}
\resizebox{\linewidth}{!}{%
\begin{tabular}{lrrr}
\toprule
compared with & held-out profiles & Composite & Quirky \\
\midrule
RUCB (pooled, non-contextual) & +51.4\% [+50.1\%, +52.7\%] & +27.0\% [+22.4\%, +31.6\%] & +36.2\% [+33.2\%, +39.5\%] \\
Pooled & +51.6\% [+48.8\%, +54.5\%] & +36.3\% [+26.9\%, +46.3\%] & +40.5\% [+34.4\%, +47.5\%] \\
CoLSTIM & +26.3\% [+24.9\%, +27.6\%] & +26.6\% [+24.0\%, +29.0\%] & +21.9\% [+19.5\%, +24.1\%] \\
COLDB & +30.5\% [+28.4\%, +32.7\%] & +22.3\% [+19.1\%, +25.5\%] & +21.8\% [+19.1\%, +24.8\%] \\
COLDB-LR & +25.4\% [+22.9\%, +27.9\%] & +26.2\% [+23.3\%, +29.3\%] & +25.1\% [+22.1\%, +28.2\%] \\
Meta-TS & +49.8\% [+47.9\%, +51.7\%] & +23.6\% [+17.2\%, +30.6\%] & +36.5\% [+32.4\%, +40.4\%] \\
HierTS & +48.5\% [+47.0\%, +50.0\%] & +20.4\% [+14.9\%, +26.2\%] & +34.6\% [+30.7\%, +38.6\%] \\
AdaTS & +48.7\% [+47.2\%, +50.2\%] & +20.6\% [+15.0\%, +26.3\%] & +34.3\% [+30.3\%, +38.3\%] \\
\textit{type-library reference} & +38.9\% [+37.1\%, +40.6\%] & +16.1\% [+10.1\%, +22.3\%] & +28.1\% [+22.4\%, +33.7\%] \\
\bottomrule
\end{tabular}
}
\caption{Second judge (TypeSafe Jev): paired reductions in excess regret of \method{} relative to each comparator, expressed as percentages of that comparator's excess regret. Brackets give 95\% bootstrap intervals (Appendix~\ref{app:uncertainty}). Table~\ref{tab:sj-main} gives the benchmark scores.}
\label{tab:sj-main-gains}
\end{table}

%% file: generated/tab_sj_topic.tex
\begin{table}[t]
\centering
\small\setlength{\tabcolsep}{3pt}
\resizebox{\linewidth}{!}{%
\begin{tabular}{lrrrr}
\toprule
method & \multicolumn{2}{c}{held-out topics} & \multicolumn{2}{c}{development} \\
 & excess & score & excess & score \\
\midrule
\textit{random pairs} & 2613.7 & 0 & 2238.3 & 0 \\
\textit{best fixed pair} & 1468.9 & 44 & 1320.1 & 41 \\
\textit{per-user linear oracle} & 279.3 & 89 & 227.8 & 90 \\
\textit{distinct-pair floor $G_T$} & 0.0 & 100 & 0.0 & 100 \\
\midrule
RUCB (pooled, non-contextual) & 1697.0 $\pm$ 5.9 & 35 & 1565.8 $\pm$ 9.0 & 30 \\
Pooled & 1840.8 $\pm$ 54.3 & 30 & 1734.6 $\pm$ 74.2 & 23 \\
CoLSTIM & 1616.2 $\pm$ 3.0 & 38 & 1383.1 $\pm$ 4.8 & 38 \\
COLDB & 1432.5 $\pm$ 13.5 & 45 & 1242.1 $\pm$ 8.2 & 45 \\
COLDB-LR & 1378.0 $\pm$ 11.8 & 47 & 1243.7 $\pm$ 11.7 & 44 \\
Meta-TS & 1602.5 $\pm$ 19.4 & 39 & 1528.0 $\pm$ 30.6 & 32 \\
HierTS & 1588.3 $\pm$ 12.2 & 39 & 1474.0 $\pm$ 16.0 & 34 \\
AdaTS & 1580.6 $\pm$ 6.7 & 40 & 1468.6 $\pm$ 10.6 & 34 \\
\method{} & \textbf{998.7 $\pm$ 4.6} & \textbf{62} & \textbf{927.8 $\pm$ 5.3} & \textbf{59} \\
\midrule
\emph{ablation:} no transfer & 1661.6 $\pm$ 4.1 & 36 & 1411.0 $\pm$ 5.2 & 37 \\
\bottomrule
\end{tabular}
}
\caption{Second judge (TypeSafe Jev): held-out-topic roster (\numJSeedsT{} seeds) and development roster (\numJSeedsD{} seeds), same layout as Table~\ref{tab:main}. The floor $G_T$, already subtracted from every entry: 843 (held-out topics), 728 (development).}
\label{tab:sj-topic}
\end{table}

%% file: sections/appendix_prompts.tex
\section{Student and tutor prompts}
\label{app:prompts}

This appendix collects the student profiles, question format, and tutor instructions used to construct the benchmark. Shaded boxes reproduce prompt components verbatim; their titles are explanatory labels.

\subsection{Hand-written student profiles}
\label{app:profile-similarity}

\begin{figure}[htbp]
\centering
\includegraphics[width=\linewidth]{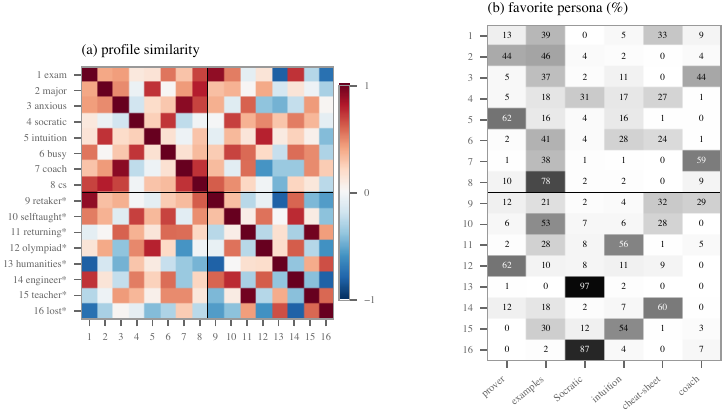}
\caption{The \numDataProfiles{} student profiles: rows 1--8 are the development profiles (measured on the development roster), rows 9--16, marked with an asterisk, the new profiles (measured on the held-out-profile roster). (a) Cosine similarity between profiles' mean question-centered Bradley--Terry scores, by position in the question chain (definition, computation, why-question) and persona. Blue and red indicate negative and positive similarity. (b) Percentage of questions on which each persona has the highest reference utility. Darker cells indicate higher percentages.}
\label{fig:dataset-quality}
\end{figure}

Profiles are compared through their mean question-centered scores at each position in the question chain, so that two profiles are similar when they prefer the same personas for the same kind of question (Figure~\ref{fig:dataset-quality}). The highest similarity between two different profiles is \numDataProfSimMax{}. On a separate bank of $24$ shared diagnostic questions, only $4$ of the $120$ pairs of profiles share a favorite persona on $70\%$ or more of the questions (at most $88\%$).

\textbf{Profile texts.}
The sixteen profiles follow verbatim, numbered as in Figure~\ref{fig:dataset-quality}. The three with a dominant favorite in Figure~\ref{fig:dataset-quality}b ask for that style explicitly: the humanities graduate student and the first-year who is lost want to be asked questions (Socratic guide, \numProfFavNewfour\% and \numProfFavNewseven\% of questions), and the software engineer wants a small example worked through (worked-examples tutor, \numProfFavDevseven\%).

\input{generated/profile_texts}

\subsection{Composite and Quirky students}
\label{app:template}

Composite and Quirky students are rendered from the following template.

\begin{promptbox}[frametitle={Composite and Quirky profile template}]
You are a student learning this material. You prefer explanations that \emph{[style]}. You like answers that are \emph{[length]}. Your preferred tone is \emph{[tone]}. When you are stuck you want the tutor to \emph{[help]}. For a definition you want \emph{[definition wish]}. For a calculation you want \emph{[calculation wish]}. For a why-question you want \emph{[why wish]}.
\end{promptbox}

The traits take the following values.
\begin{itemize}
\small
\item Style: ``start from precise definitions and justify claims''; ``work through a concrete numerical example first''; ``explain with an analogy or a mental picture''; ``mix a short definition with one small example''.
\item Length: ``as short as possible''; ``of ordinary lecture length''; ``patient and unhurried''.
\item Tone: ``formal and impersonal''; ``warm and encouraging''.
\item Help when stuck: ``tell you the answer directly''; ``guide you with leading questions''.
\item Definition wish: ``a compact definition you can quote''; ``an analogy before any symbols''; ``a precise statement with hypotheses''; ``a gentle numbered introduction''.
\item Calculation wish: ``one fully worked numerical example''; ``careful algebraic steps''; ``a compressed procedure''; ``numbered steps you can imitate''.
\item Why-question wish: ``a real proof or proof sketch''; ``the idea in plain words and a picture''; ``one sentence of justification''; ``questions that let you reconstruct the argument''.
\end{itemize}
A Composite student draws style, length, tone and help with fixed probabilities (style $0.2/0.3/0.2/0.3$, length $0.25/0.5/0.25$, tone $0.5/0.5$, help $0.7/0.3$) and the three wishes uniformly; the $350$ arrivals are distinct combinations drawn without replacement among the $3{,}072$. A Quirky student draws every trait uniformly and appends two distinct quirks from: ``You like a brief application to sport if it fits.'', ``You dislike numbered lists.'', ``You want a one-line historical aside when it is natural.'', ``You prefer British spelling.'', ``You want the answer to mention a programming connection when there is one.'', ``You dislike rhetorical questions.'', ``You like a tiny diagram described in words.'', ``You want units and dimensions mentioned on numerical examples.'', ``You dislike exclamation marks.'', ``You want a counter-example when the tutor states a rule.'' Every Quirky arrival has a distinct combination of traits and quirks.

\subsection{Question construction}
\label{app:cues}

Each session covers four topics with two chains of three questions each: a definition question, a computation or numerical example, and a why-question. The second chain of a topic asks about a different aspect of it. The session asks the four first chains, then the four second chains.

Every question ends with one of the following $13$ statements about the student's situation or needs. We draw one uniformly from the statements for that question type, using a fixed seed.

\begin{promptbox}[frametitle={Statements in definition questions}]
\begin{itemize}
\small
\item ``This vocabulary is new to me and a formula with no picture is not landing.''
\item ``I have never seen one written out with actual numbers.''
\item ``I need a line I can quote in an exam.''
\item ``I have heard the story version already; I want the real statement.''
\end{itemize}
\end{promptbox}

\begin{promptbox}[frametitle={Statements in computation questions}]
\begin{itemize}
\small
\item ``I already know the rule exists; I just need the working.''
\item ``Show me one so I can copy the method.''
\item ``I need a write-up I can hand in.''
\item ``I need a number I can use, not a lecture.''
\item ``I got a different result twice and I cannot see where I go wrong.''
\end{itemize}
\end{promptbox}

\begin{promptbox}[frametitle={Statements in why-questions}]
\begin{itemize}
\small
\item ``I can use it but I do not see why it is allowed.''
\item ``I believed this in school; I want to know why it is true.''
\item ``I want the actual argument, not another walkthrough.''
\item ``I already believe the simple case; I want to know why it holds in general.''
\end{itemize}
\end{promptbox}

\subsection{Tutor generation prompts}
\label{app:tutor-prompts}

For each question, the system message contains one persona prompt below, then a blank line, then the shared answer instruction. The user message contains the question. We sample one answer per persona and question at temperature $1$. These are the full tutor prompts used for every roster. Each question is answered independently, without earlier questions or answers in the generation context.

\begin{promptbox}[frametitle={Shared answer instruction}]
Answer the student's question below. Write between 100 and 180 words (between 40 and 90 words if your style is the compact cheat-sheet). Plain text with light LaTeX-free notation; no headings. The student's message may say what they already know or what they need. Take it into account, including for the length of your answer if needed (you may then leave the word range above), while keeping your own teaching style.
\end{promptbox}

\begin{promptbox}[frametitle={Rigorous prover}]
You are a mathematics tutor who teaches through rigor. Always start from precise definitions, state results as claims and prove them or sketch the proof, use standard notation, and point out where hypotheses are needed. Avoid analogies and avoid chatty language. Do not ask the student questions.
\end{promptbox}

\begin{promptbox}[frametitle={Worked-examples tutor}]
You are a tutor who teaches through concrete worked examples. Answer by immediately working through one or two specific numerical or concrete examples in full detail, then state the general pattern briefly at the end. Keep notation light. Do not ask the student questions.
\end{promptbox}

\begin{promptbox}[frametitle={Socratic guide}]
You are a Socratic tutor. Do not give the full answer directly. Instead, guide the student with a short sequence of leading questions and small hints, each building on the previous, so that they can reach the answer themselves. Give the key idea only at the end, in one sentence.
\end{promptbox}

\begin{promptbox}[frametitle={Intuition-and-analogy tutor}]
You are a tutor who teaches through intuition, analogies and big-picture explanations. Explain the idea in plain language with a vivid analogy or mental picture, say why the idea matters, and keep formulas to a minimum. Do not ask the student questions.
\end{promptbox}

\begin{promptbox}[frametitle={Concise cheat-sheet}]
You are a tutor who answers in the most compact way possible: a short bullet-point summary with the key definition, the key formula or rule, and one line on when it applies. No analogies, no questions, no encouragement, no examples beyond one line. Aim for under 90 words.
\end{promptbox}

\begin{promptbox}[frametitle={Patient step-by-step coach}]
You are a patient, encouraging tutor. Answer with a clearly numbered step-by-step procedure the student can follow, explain the purpose of each step in simple words, warn about common mistakes, and end with a short encouraging remark. Do not ask the student questions.
\end{promptbox}

\subsection{Rater query}
\label{app:rater-prompt}

The rater receives the student's question as its only input, under the key \texttt{student\_message}, and answers the query below. Its six options are the persona style descriptions, keyed by persona.

\begin{promptbox}[frametitle={Rater query}]
Which style of answer would serve what the student asks for in \textasciigrave student\_message\textasciigrave{} worst?
\end{promptbox}

\begin{promptbox}[frametitle={Style descriptions (the query's options)}]
\textbf{proof}: A rigorous answer: it starts from precise definitions, states the result as a claim and proves it or sketches the proof, with standard notation and the hypotheses that are needed; no analogies.\\
\textbf{examples}: A worked-example answer: it immediately works through one or two specific numerical or concrete examples in full detail, then states the general pattern briefly at the end, with light notation.\\
\textbf{socratic}: A Socratic answer: it does not give the full answer, but guides the student with a short sequence of leading questions and small hints so they reach the answer themselves, giving the key idea only at the end.\\
\textbf{intuition}: An intuitive answer: it explains the idea in plain language with a vivid analogy or mental picture, says why the idea matters, and keeps formulas to a minimum.\\
\textbf{concise}: A minimal answer: a short bullet-point summary with the key definition, the key formula or rule, and one line on when it applies, in under 90 words.\\
\textbf{steps}: A step-by-step answer: a clearly numbered procedure the student can follow, the purpose of each step in simple words, warnings about common mistakes, and a short encouraging remark at the end.
\end{promptbox}

%% file: generated/profile_texts.tex
\begin{promptbox}[frametitle={Student profile 1. exam-focused student (development)}]
You are twenty, it is week 12, and you have three midterms. You want the official line you can quote, because that is what the mark scheme pays. Last sitting a word you had never seen made a crib useless until you had one example of the object. If you already know the drill you just want the shortest working. Stories and being asked to discover the answer make you anxious about the clock. You are trying to survive the exam, not become a mathematician.
\end{promptbox}

\begin{promptbox}[frametitle={Student profile 2. mathematics major (development)}]
You are a mathematics major, but last year a formal lecture on generating functions left you blank until a friend computed three small examples on a napkin. You still take pride in a real argument, and notation does not scare you once you know what the object is. Vague cheerleading does. When a result feels like a trick you want to see why anyone would believe it before you will read the write-up. You have stayed up finishing problem sets with friends who write every algebra step, and on those nights a careful calculation feels more honest than a sweeping claim. You still dislike being encouraged. You want to know what is actually true. A warm explanation that never says why it is allowed still feels empty.
\end{promptbox}

\begin{promptbox}[frametitle={Student profile 3. anxious student (development)}]
You are eighteen, first term at university, and you still half-expect someone to tell you the offer was a mistake. You reread the same paragraph four times and then close the laptop because your chest is tight. You do not want a pretty story and you do not want to be left alone with a riddle. You want a voice that stays with you, names what just happened in ordinary words, and does not make a joke of your panic. If they skip a line you assume it is because you are slow. Silence in a tutorial feels like drowning. You will take a slightly long, careful walkthrough from someone who sounds kind. If they ask why a rule is allowed you still want those ordinary words for what just happened, not a flourish. You will not take a flourish, and you will not take being put on the spot before you have even copied the heading.
\end{promptbox}

\begin{promptbox}[frametitle={Student profile 4. self-reconstructing student (development)}]
You are the kind of student who pauses lectures to try the next step yourself. Being handed a finished solution feels like someone spoiling a puzzle, and you do not remember things you did not reconstruct. Dense crib sheets go in a drawer. Sunday night before a deadline you have a short fuse: if you already understand the idea and you only need a number, being quizzed feels like contempt, and you want the shortest working on the page so you can move. You will sit through a careful calculation when you are truly stuck and the hints have gone in a circle. Reconstruction is the default, not a religion.
\end{promptbox}

\begin{promptbox}[frametitle={Student profile 5. student who reads for the ideas (development)}]
You are taking this because the ideas are beautiful, not because you need the credit. You read popular maths books for fun. A tutor who only pushes symbols feels like they missed why anyone would care. You like metaphors, history, and a sense of what is actually going on. You still want to leave with a picture of why the idea is beautiful. You once trusted a pretty story that turned out to be wrong, and you still wince. When you need to know whether a claim is actually true, or why a rule is allowed, a metaphor is not enough to bet on --- you want the real argument, even if it is dry. Formal write-ups are welcome when they illuminate. You have little patience for classroom cheerleading or for notes that look like an exam crib. A pretty story is no substitute for the working either.
\end{promptbox}

\begin{promptbox}[frametitle={Student profile 6. busy student (development)}]
You are thirty-four, you work full time, and you are watching a lecture on your phone between meetings. You already have a degree in something else, so a lot of this is foreign vocabulary sitting on top of skills you do have. You are not fragile and you are not here for pedagogy experiments. If it looks like something you already use at work you want the usable bit and then you close the laptop. If it is a word you have never said in a meeting, a cheat sheet you cannot interpret is a waste; you want one picture that tells you what you are looking at, then the formula. You have sat through a proof of a word you could not yet picture, and you will not do that again. Warmth still reads as padding. You will not do extra exercises for the vibe.
\end{promptbox}

\begin{promptbox}[frametitle={Student profile 7. student who wants a coach (development)}]
You learned algebra with a teacher who named every step and the usual trap. That is still how you learn when the page looks like something you have done before. A wall of notation or a clever analogy with no scaffolding and you are gone. When the words are new, even a careful walkthrough does not land until someone shows you one tiny worked example. You would rather a slightly long explanation you can follow than a short one you have to decode. Being quizzed before you have a path still panics you.
\end{promptbox}

\begin{promptbox}[frametitle={Student profile 8. software engineer (development)}]
You write software for a living and you wandered into this course because a model you shipped started using linear algebra you only half remembered. You trust a walkthrough that looks like a notebook cell: pick a small input, run it, print the intermediates, see the shape change. Elegant arguments that never touch a number make you suspicious, as if someone skipped the unit test. You have shipped bugs that only showed up once, so you will not leave a session without knowing what is supposed to stay true. You do not want to be soothed and you do not want to be quizzed in public. If the tutor talks like a novelist you mute them. You mute anyone who will not pick a small input and run it. Show the object moving.
\end{promptbox}

\begin{promptbox}[frametitle={Student profile 9. student retaking the course (test)}]
You are retaking this course after failing it last year. You are not stupid, you froze. What helps you now is being told exactly what to write and why it is right, in that order; anything that wanders makes you feel the clock again. You will read a longer answer if every line earns its place, and you have stopped trusting encouragement that is not attached to a concrete step.
\end{promptbox}

\begin{promptbox}[frametitle={Student profile 10. self-taught programmer (test)}]
You taught yourself to program from documentation and forum answers, and you are filling in the mathematics that your work now needs. You read quickly, skip what you already know, and get irritated by explanations that start from zero. What convinces you is a small case you can run in your head, followed by the general statement. You do not need reassurance; you need the idea to compile. You would rather see one line of the general rule after the case than three paragraphs before it.
\end{promptbox}

\begin{promptbox}[frametitle={Student profile 11. adult returning to study (test)}]
You are forty-one, back in a classroom after fifteen years in a job that used none of this. Vocabulary is the wall: a term you have never heard turns the rest of the paragraph to noise. You want one picture or comparison that tells you what kind of thing you are looking at, then the formula, then a plain example. You are patient and you will reread, but you resent being talked down to.
\end{promptbox}

\begin{promptbox}[frametitle={Student profile 12. olympiad student (test)}]
You are seventeen and you do competition mathematics for fun. You want the real reason a thing is true, stated cleanly, and you are bored by procedures you could have found yourself. A clever short argument delights you; a numbered recipe with reassuring remarks feels like being handed the answer key. You do not mind notation, and you notice when a hypothesis is quietly missing. Being told to try it yourself is fine when the hint is sharp; a vague nudge just wastes your evening.
\end{promptbox}

\begin{promptbox}[frametitle={Student profile 13. humanities graduate student (test)}]
You are a graduate student in history who needs this material for a methods requirement. You learn by being asked what an idea means, and by answering in a sentence of your own before anyone writes a formula. A metaphor you cannot rephrase is decoration. A finished write-up you only read does not stay with you. You like a question you can actually answer, and you shut down if they ask you to compute before anyone has shown you how. You are not looking for a numbered recipe or a picture. You are looking for the moment you can say the point yourself.
\end{promptbox}

\begin{promptbox}[frametitle={Student profile 14. engineer with a deadline (test)}]
You are an engineer with a deadline. You will not touch a cheat sheet, and you will not sit through a pile of examples. You want one sentence that says what the object is and why the operation is legal, then the arithmetic of one measurement you would actually take, and then you are done. A second illustration feels like stalling. A proof that never reaches a quantity can wait. Encouragement can wait. If you cannot point to the permission and to one number that changed, you are not finished for today.
\end{promptbox}

\begin{promptbox}[frametitle={Student profile 15. teacher preparing a class (test)}]
You are a secondary-school teacher preparing to explain this to a class next week. You already know the mechanics; what you are looking for is the explanation that would make a sixteen-year-old nod: a picture, a small example, the one sentence that says why. You judge an answer by whether you could reuse it in front of a room, and you dislike rigour that only a specialist could follow. Long careful walkthroughs are fine when they hand you a structure you could copy onto a board.
\end{promptbox}

\begin{promptbox}[frametitle={Student profile 16. first-year who is lost (test)}]
You are in your first term, you did not follow the lecture, and you are embarrassed to ask. You need every step named in ordinary words, slowly. Watching a finished example does not help you: you need to be asked for the next line before anyone writes it. A wall of symbols makes you close the page. Warmth helps you keep reading. A crib you cannot walk through line by line feels like another room you are not allowed into. One path, named as you go, is what makes a definition stick.
\end{promptbox}

%% file: sections/appendix_results.tex
\section{Additional Results}
\label{app:results}

\subsection{TutorDuel results}
\label{app:results-tutorduel}
\suppressfloats[t]

\textbf{Per profile.}
The held-out-profile roster has only eight profiles, so its total depends on which ones it contains. Table~\ref{tab:profiles} splits it by profile. \method{} scores from \numUMethodProfileWorst{} to \numUMethodProfileBest{} across profiles and exceeds \numUProfileOtherName{} on every profile. Resampling the profiles and the seeds after the fact puts its score between \numUProfileBootLo{} and \numUProfileBootHi{} (95\%), against \numUMethodGap{} for these eight profiles.

\input{generated/tab_profiles}

\textbf{Other rosters of the same kind.}
The standard errors in Table~\ref{tab:main} treat each roster as fixed. To see how the results change with another roster of the same kind, every seed draws a new roster and reruns the policies on it (Table~\ref{tab:resampled}): $350$ Composite or Quirky students drawn with replacement from the same pool, or eight held-out profiles drawn with replacement. On Composite and Quirky, \method{} scores \numCRsMethodGap{} (\numCRsMethodGapLo{} to \numCRsMethodGapHi{}) and \numIRsMethodGap{} (\numIRsMethodGapLo{} to \numIRsMethodGapHi{}), close to its score on the actual rosters. On every redrawn roster, \method{} has lower excess regret than each comparison method and the no-transfer ablation; its smallest average lead is over COLDB, \numCRsOverColdb{} and \numIRsOverColdb{} points, the same as on the actual rosters. A redrawn held-out-profile roster draws its eight profiles with replacement, so it has on average \numURsDistinct{} distinct profiles and more users per profile than the actual roster. Methods that share information across users then score higher than on the actual roster, and \method{} scores \numURsMethodGap{}.

\input{generated/tab_resampled}

\textbf{Paired comparisons.}
Table~\ref{tab:main-gains} pairs \method{} with each comparison method on the three evaluation rosters (Section~\ref{sec:res-main}).

\input{generated/tab_main_full}
\input{generated/tab_main_gains}

\textbf{Held-out-topic roster and development roster.}
The held-out-topic roster keeps the eight development profiles and changes only the questions, so it separates the effect of new topics from that of new students. Table~\ref{tab:topic} is its counterpart of Table~\ref{tab:main}, together with the development roster, and Table~\ref{tab:topic-gains} the counterpart of Table~\ref{tab:main-gains}. Because the development roster was also used for design, its numbers are descriptive. The fixed-pair references are chosen on the development roster for the held-out-topic roster and cross-fitted across arrivals on the development roster itself (chosen on one half of the users of each profile, paid on the other). There \method{} scores \numTMethodGap{}, and a reference that knows each development profile's choice probabilities and utilities, and recognizes which profile a user follows, scores \numTTypeLibGap{}. With new students as well, \method{} scores \numTabMainMethodGapLo{} to \numTabMainMethodGapHi{}, while this reference falls to \numTypeLibMainLo{} to \numTypeLibMainHi{}. What is known about the development profiles does not carry over to new students, so a method must learn the new population online.

\input{generated/tab_topic}
\input{generated/tab_topic_gains}

\input{generated/tab_phases}

\clearpage
\subsection{Numerical simulator results}
\label{app:results-simulator}

Table~\ref{tab:sim} reports results from the numerical simulator of Section~\ref{sec:exp-simulator}, using $300$ users, $20$ choices per user, and click-consistency parameters $\beta\in\{4,12\}$. These settings cover the finite-type, Gaussian, low-rank, and null populations described in Appendix~\ref{app:simulator}. The population oracle starts each user from the true population prior and provides an empirical reference in this environment. \method{} gains most where users form a few types, removing \numSimRemovedTwoTypesLo{} to \numSimRemovedTwoTypesHi\% of the no-transfer ablation's excess regret with two types, and less as users become harder to group: \numSimRemovedEightTypesLo{} to \numSimRemovedEightTypesHi\% with eight types and \numSimRemovedGaussLo{} to \numSimRemovedGaussHi\% for a Gaussian cloud. No comparison method has lower excess regret than \method{} in any of the \numSimAnchorCells{} structured settings. On the null population, every comparison method that shares information across users raises excess regret over no transfer, by \numSimNullSharersLo{} to \numSimNullSharersHi{}, while \method{} stays within noise of it.

\input{generated/tab_sim}

\textbf{Sensitivity to click consistency and session length.}
Across the simulator's $b\times\beta$ grid ($16$ seeds per cell), \method{} has significantly lower excess regret than the no-transfer ablation in \numSimGridBetter{} of the \numSimGridCells{} settings with a structured population, significantly higher in none, and unresolved differences in the others (paired $t$-test, Benjamini--Hochberg at $q=0.05$). The mean reduction, as a percentage of the ablation's excess regret, rises from \numSimTrendBetaLo\% at $\beta=\numSimTrendBetaFirst$ to \numSimTrendBetaHi\% at $\beta=\numSimTrendBetaLast$ and from \numSimTrendBLo\% at $b=\numSimTrendBFirst$ to \numSimTrendBHi\% at $b=\numSimTrendBLast$. On the null population, \method{} has significantly higher excess regret in \numSimNullWorse{} of \numSimNullCells{} settings (mean reduction \numSimNullMeanShare\%).

\subsection{Comparison methods}
\label{app:comparison-checks}

We compare learning policies on the same user and question sequences, with one binary choice per round and the same number of rounds. Each policy observes choices only on the pairs it selects. In \benchmark{}, contextual methods receive the same persona indicators and rater scores; \method{} learns its recalibration from completed sessions. Policies start each roster without learned user preferences and do not receive student profiles or evaluation utilities. The subsections below describe their adaptations, tuning, and additional checks.

\subsubsection{COLDB and COLDB-LR}
COLDB~\citep{wang2025online} pools comparisons within groups of users. Our sequential implementation connects each new user to all earlier users and removes links when their individual preference estimates differ by more than a distance threshold, which decreases as they make more choices. To support grouping from short sessions, we also evaluate COLDB-LR. It keeps COLDB's pooled logistic model and pair-selection rule but changes group membership using a likelihood-ratio criterion. Each user starts alone and, before its first choice, uses the model pooled over earlier users. After user $i$ has made $n_i$ choices, it joins the group that best predicts those choices, provided that the group's log likelihood is no more than $\mathrm{tol}\,(d/2)[1+\log(1+n_i)]$ below that of its own fitted model. Group estimates exclude that user's data. If no group qualifies, the user is isolated. We tune $\mathrm{tol}$ as in Table~\ref{tab:tuning}.

In the numerical simulator, COLDB's excess regret at $b=20$ is close to Pooled's (Table~\ref{tab:sim}). We therefore also checked longer sessions. On the two-type, $90^\circ$ population with $\beta=4$ ($16$ seeds), COLDB with distance and exploration scales both set to $0.03$ produced an average of $\numColdbLongClusters{}$ clusters at $b=80$, with excess regret $\numColdbLongSplit{}$ versus $\numColdbLongPooled{}$ for Pooled. With both scales set to $1$, it retained one cluster at $b=160$. COLDB-LR with tolerance $0.5$ and exploration scale $0.3$ produced an average of $\numColdbLrShortClusters{}$ clusters at $b=40$.

\subsubsection{Hyperparameter tuning}
Table~\ref{tab:tuning} lists the grids searched for CoLSTIM, COLDB, COLDB-LR, RUCB, Meta-TS, HierTS, and AdaTS, and for the margin of the latent-bandit UCB ablation of \method{}, with the chosen configurations. Each configuration is run on $16$ tuning seeds separate from those used for benchmark evaluation, on the development roster for the tutoring benchmark and on the $16$ structured $b=20$ anchors for the simulator, and scored by its excess regret relative to no transfer averaged over worlds. The chosen configuration of each method is then used for every reported comparison in that environment. When the best configuration lies on the edge of a grid, the grid is extended along that axis and the selection is redone before any evaluation result of the method is read: Meta-TS's grid covers \numTuneMetaTsConfigs{} configurations after this step. Three selections stay on an edge: RUCB's $\alpha=0.51$, since its analysis requires $\alpha>1/2$; AdaTS's smallest prior scale, which approaches pooling all users into one model; and CoLSTIM's smallest noise scale on the simulator, where the perturbation does not help and smaller scales approach the unperturbed leader (we extended this axis once, from $0.1$ down to $0.01$, and the ratio kept falling).

\input{generated/tab_tuning}

\subsubsection{Settings chosen on the evaluation rosters}
The comparison methods were tuned on the development roster, so their settings may suit the development students better than the new ones. To measure how much this costs them, we reran every configuration of the grids of Table~\ref{tab:tuning} on the three evaluation rosters and kept, for each method and roster, the configuration with the lowest excess regret on that roster (Table~\ref{tab:tuning-oracle}). These configurations are chosen on the evaluation data, so their scores are optimistic. Even so, the best configuration of any comparison method removes \numUOracleBestScore\% of random pairs' excess regret on the held-out profiles, where \method{} removes \numUOracleMethodScore\%. On Composite and Quirky the best comparison methods reach \numCOracleBestScore\% and \numIOracleBestScore\%, against \numCOracleMethodScore\% and \numIOracleMethodScore\% for \method{}. COLDB gains most from the second chance: its development setting removes \numUOracleColdbPaper\% on the held-out profiles and its best setting there removes \numUOracleColdbBest\%. The three methods that learn a population mean, Meta-TS, HierTS, and AdaTS, keep more excess regret than the no-transfer ablation on the held-out profiles whatever their prior variances. We also looked beyond the edge of RUCB's grid. A larger $\alpha$ makes RUCB explore more, and the extra exploration raised its excess regret on every roster, so the smallest admissible value remains its best setting. Table~\ref{tab:ts-pair} reports the check of the double-sample pair rule for Meta-TS, HierTS, and AdaTS described in Section~\ref{sec:exp-policies}.

\input{generated/tab_tuning_oracle}
\input{generated/tab_ts_pair}

\subsection{Sensitivity to click sharpness and session length}
\label{app:click-sensitivity}
\begin{figure}[!htbp]
\centering
\includegraphics[width=\textwidth]{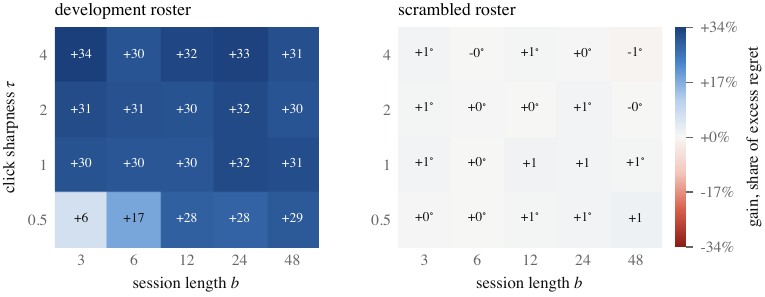}
\caption{Paired gain of \method{} over the no-transfer ablation as a share of the ablation's excess regret, by click sharpness $\tau$ and session length $b$ ($32$ seeds per cell; $\tau\neq1$ uses the Bradley--Terry-consistent click table), on the development roster and on its scrambled version. The scrambled panel checks for increased regret when persona labels are permuted independently across users. A small circle marks cells not resolved at Benjamini--Hochberg level $0.05$. A session of length $b<24$ asks its first $b$ questions; $b=48$ repeats the private pack.}
\label{fig:llm-maps}
\end{figure}

Figure~\ref{fig:llm-maps} varies the click sharpness $\tau$ and the session length $b$ on the development roster and reports the paired gain of \method{} over the no-transfer ablation as a share of the ablation's excess regret. Logits are multiplied by $\tau$ (for $\tau\neq1$ the logits are the differences of the personas' utilities, so that choices follow the Bradley--Terry model, instead of the stored choice probabilities, and both learners of \method{} use $\beta=\tau$), a session of length $b<24$ asks its first $b$ questions, and longer sessions repeat the private pack. \method{} has significantly lower excess regret in all tested settings, with gains between \numDevGridShareLo\% and \numDevGridShareHi\% of the ablation's excess. The gain is between \numDevGridRestLo\% and \numDevGridRestHi\% in every cell except the two with the softest clicks and the shortest sessions ($\tau=0.5$ with $b=3$ or $6$), where the gains are smaller. On the scrambled roster, with persona labels permuted independently for each user, gains range from \numScrambledGridLo\% to \numScrambledGridHi\%. \method{} is significantly worse in no setting and significantly better in \numScrambledBetter{}.

\begin{figure}[!htbp]
\centering
\includegraphics[width=0.6\textwidth]{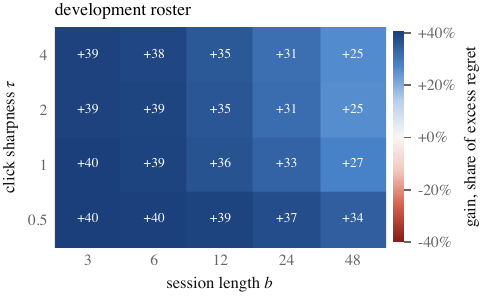}
\caption{Development roster: paired gain of the type-library reference over no transfer as a share of the latter's excess regret, by click sharpness $\tau$ and session length $b$ ($32$ seeds per cell). This is a comparison of references, not a sensitivity analysis of \method{}.}
\label{fig:llm-maps-typelib}
\end{figure}

\begin{samepage}
\subsubsection{Reference-only sensitivity map}
Figure~\ref{fig:llm-maps-typelib} varies click sharpness and session length for the type-library reference against no transfer on the development roster. With the development profiles' tables, the reference has significantly lower excess regret than no transfer in all tested settings; on this roster the tables describe every user. This figure compares references and does not evaluate \method{}.\par
\end{samepage}

\subsection{Diagnostic references}
Table~\ref{tab:references} compares three diagnostic references. The known-profile pair nearly matches the best fixed pair per user on the rosters with recurring profiles. The type-library reference has a lower score than the per-user fixed-pair reference on every roster.

\input{generated/tab_references}

%% file: generated/tab_profiles.tex
\begin{table}[!htbp]
\centering
\small
\begin{tabular}{lrrr}
\toprule
profile & users ($n$) & \method{} & CoLSTIM \\
\midrule
retaking the course & 44 & 29 & 25 \\
self-taught programmer & 44 & 44 & 26 \\
returning adult & 44 & 56 & 33 \\
olympiad student & 44 & 36 & 30 \\
humanities graduate & 44 & 61 & 45 \\
engineer with a deadline & 44 & 42 & 25 \\
teacher preparing a class & 43 & 59 & 36 \\
lost first-year & 43 & 55 & 44 \\
\bottomrule
\end{tabular}
\caption{Held-out-profile roster, by profile: benchmark score on that profile's users (share of random pairs' excess regret removed), mean over \numSeedsU{} seeds, from the runs of Table~\ref{tab:main} split after the fact. Higher is better. \numUProfileOtherName{} is the comparison method with the lowest excess regret on the full held-out-profile roster.}
\label{tab:profiles}
\end{table}

%% file: generated/tab_resampled.tex
\begin{table}[!htbp]
\centering
\small\setlength{\tabcolsep}{4pt}
\resizebox{\linewidth}{!}{%
\begin{tabular}{lrrr}
\toprule
policy & held-out profiles & Composite & Quirky \\
\midrule
\method{} & 55 [45, 64] & 45 [42, 47] & 46 [43, 49] \\
COLDB & 33 [8, 54] & 38 [32, 43] & 38 [26, 45] \\
CoLSTIM & 34 [29, 39] & 28 [26, 30] & 29 [26, 31] \\
Meta-TS & 32 [12, 53] & 34 [27, 40] & 36 [27, 44] \\
HierTS & 33 [15, 55] & 36 [29, 41] & 38 [29, 44] \\
\emph{ablation:} no transfer & 31 [27, 35] & 27 [25, 29] & 27 [25, 30] \\
\midrule
\method{} minus COLDB & +21 [+4, +42] & +6 [+2, +14] & +8 [+2, +20] \\
\method{} minus CoLSTIM & +21 [+12, +29] & +17 [+14, +20] & +18 [+14, +22] \\
\method{} minus Meta-TS & +23 [+8, +36] & +10 [+5, +17] & +10 [+4, +19] \\
\method{} minus HierTS & +22 [+7, +37] & +9 [+3, +17] & +9 [+3, +16] \\
\method{} minus no transfer & +24 [+15, +30] & +17 [+15, +20] & +19 [+15, +22] \\
\bottomrule
\end{tabular}
}
\caption{Benchmark score over redrawn rosters: every seed draws a new roster of the same kind (the eight held-out profiles drawn with replacement; $350$ Composite or Quirky students drawn with replacement), and the policies are rerun on it. Mean over \numURsSeeds{} seeds and, in brackets, the 2.5th and 97.5th percentiles of the per-seed values, which describe the spread of a new roster of that kind (students, arrival order and clicks together). The last rows report paired differences in score points, computed for methods run on the same redrawn roster and seed. Their ranges show how the advantage varies across rosters.}
\label{tab:resampled}
\end{table}

%% file: generated/tab_main_full.tex
\begin{table}[t]
\centering
\small\setlength{\tabcolsep}{3pt}
\resizebox{\linewidth}{!}{%
\begin{tabular}{lrrrrrr}
\toprule
method & \multicolumn{2}{c}{held-out profiles} & \multicolumn{2}{c}{Composite} & \multicolumn{2}{c}{Quirky} \\
 & excess & score & excess & score & excess & score \\
\midrule
\textit{random pairs} & 2317.3 & 0 & 2276.1 & 0 & 2305.5 & 0 \\
\textit{best fixed pair} & 2197.5 & 5 & 1900.0 & 17 & 1790.3 & 22 \\
\textit{per-user linear oracle} & 356.8 & 85 & 427.8 & 81 & 411.8 & 82 \\
\textit{distinct-pair floor $G_T$} & 0.0 & 100 & 0.0 & 100 & 0.0 & 100 \\
\midrule
RUCB (pooled, non-contextual) & 1918.5 $\pm$ 5.9 & 17 & 1557.1 $\pm$ 8.1 & 32 & 1528.8 $\pm$ 7.0 & 34 \\
Pooled & 1919.5 $\pm$ 30.6 & 17 & 1565.7 $\pm$ 33.4 & 31 & 1552.8 $\pm$ 49.6 & 33 \\
CoLSTIM & 1546.0 $\pm$ 3.6 & 33 & 1636.2 $\pm$ 4.8 & 28 & 1649.3 $\pm$ 4.2 & 28 \\
COLDB & 1801.4 $\pm$ 22.1 & 22 & 1407.2 $\pm$ 11.7 & 38 & 1428.0 $\pm$ 28.2 & 38 \\
COLDB-LR & 1812.5 $\pm$ 7.2 & 22 & 1507.4 $\pm$ 14.0 & 34 & 1478.9 $\pm$ 12.3 & 36 \\
Meta-TS & 1794.5 $\pm$ 5.0 & 23 & 1494.4 $\pm$ 12.4 & 34 & 1491.6 $\pm$ 17.3 & 35 \\
HierTS & 1773.4 $\pm$ 4.4 & 23 & 1426.2 $\pm$ 7.5 & 37 & 1438.1 $\pm$ 12.6 & 38 \\
AdaTS & 1792.4 $\pm$ 10.0 & 23 & 1442.7 $\pm$ 10.7 & 37 & 1457.2 $\pm$ 18.6 & 37 \\
\method{} & \textbf{1197.8 $\pm$ 5.1} & \textbf{48} & \textbf{1276.3 $\pm$ 6.1} & \textbf{44} & \textbf{1250.2 $\pm$ 5.4} & \textbf{46} \\
\midrule
\emph{ablation:} no transfer & 1617.2 $\pm$ 2.9 & 30 & 1658.7 $\pm$ 5.5 & 27 & 1670.9 $\pm$ 4.6 & 28 \\
\bottomrule
\end{tabular}
}
\caption{Every comparison method on the three evaluation rosters, same layout as Table~\ref{tab:main}. The floor $G_T$, already subtracted from every entry: 779 (held-out profiles), 699 (Composite), 679 (Quirky).}
\label{tab:main-full}
\end{table}

%% file: generated/tab_main_gains.tex
\begin{table}[t]
\centering
\small\setlength{\tabcolsep}{4pt}
\resizebox{\linewidth}{!}{%
\begin{tabular}{lrrr}
\toprule
compared with & held-out profiles & Composite & Quirky \\
\midrule
RUCB (pooled, non-contextual) & +37.6\% [+36.3\%, +38.8\%] & +18.0\% [+15.4\%, +20.9\%] & +18.2\% [+15.4\%, +20.9\%] \\
Pooled & +37.6\% [+34.3\%, +41.1\%] & +18.5\% [+14.1\%, +22.9\%] & +19.5\% [+13.0\%, +26.1\%] \\
CoLSTIM & +22.5\% [+21.2\%, +23.8\%] & +22.0\% [+19.9\%, +24.1\%] & +24.2\% [+22.0\%, +26.5\%] \\
COLDB & +33.5\% [+31.0\%, +36.3\%] & +9.3\% [+5.9\%, +12.7\%] & +12.5\% [+7.7\%, +17.4\%] \\
COLDB-LR & +33.9\% [+32.5\%, +35.3\%] & +15.3\% [+12.6\%, +18.1\%] & +15.5\% [+12.8\%, +18.2\%] \\
Meta-TS & +33.3\% [+31.9\%, +34.5\%] & +14.6\% [+11.5\%, +17.9\%] & +16.2\% [+12.6\%, +19.4\%] \\
HierTS & +32.5\% [+31.1\%, +33.7\%] & +10.5\% [+6.9\%, +14.3\%] & +13.1\% [+9.8\%, +16.2\%] \\
AdaTS & +33.2\% [+31.6\%, +34.8\%] & +11.5\% [+8.1\%, +15.2\%] & +14.2\% [+10.4\%, +17.7\%] \\
\textit{type-library reference} & +32.9\% [+31.5\%, +34.3\%] & +16.9\% [+13.3\%, +20.6\%] & +19.3\% [+15.3\%, +23.2\%] \\
\bottomrule
\end{tabular}
}
\caption{Paired comparison of \method{} with each comparison method: how much lower its excess regret is, as a share of that method's excess (positive: \method{} is better). The 95\% interval resamples seeds and users after the fact, so it is conditional on these students (Appendix~\ref{app:uncertainty}).}
\label{tab:main-gains}
\end{table}

%% file: generated/tab_topic.tex
\begin{table}[t]
\centering
\small\setlength{\tabcolsep}{3pt}
\resizebox{\linewidth}{!}{%
\begin{tabular}{lrrrr}
\toprule
method & \multicolumn{2}{c}{held-out topics} & \multicolumn{2}{c}{development} \\
 & excess & score & excess & score \\
\midrule
\textit{random pairs} & 2341.4 & 0 & 2000.9 & 0 \\
\textit{best fixed pair} & 1315.0 & 44 & 1173.9 & 41 \\
\textit{per-user linear oracle} & 364.6 & 84 & 307.0 & 85 \\
\textit{distinct-pair floor $G_T$} & 0.0 & 100 & 0.0 & 100 \\
\midrule
RUCB (pooled, non-contextual) & 1389.1 $\pm$ 4.3 & 41 & 1257.6 $\pm$ 6.6 & 37 \\
Pooled & 1429.1 $\pm$ 18.9 & 39 & 1385.5 $\pm$ 57.8 & 31 \\
CoLSTIM & 1632.1 $\pm$ 3.1 & 30 & 1381.8 $\pm$ 4.4 & 31 \\
COLDB & 1323.7 $\pm$ 6.2 & 43 & 1189.4 $\pm$ 7.5 & 41 \\
COLDB-LR & 1340.6 $\pm$ 3.1 & 43 & 1205.2 $\pm$ 5.2 & 40 \\
Meta-TS & 1364.6 $\pm$ 5.8 & 42 & 1226.9 $\pm$ 7.0 & 39 \\
HierTS & 1339.5 $\pm$ 3.2 & 43 & 1200.8 $\pm$ 3.5 & 40 \\
AdaTS & 1367.4 $\pm$ 15.5 & 42 & 1210.4 $\pm$ 6.5 & 40 \\
\method{} & \textbf{1087.4 $\pm$ 3.7} & \textbf{54} & \textbf{971.8 $\pm$ 4.7} & \textbf{51} \\
\midrule
\emph{ablation:} no transfer & 1673.1 $\pm$ 3.1 & 29 & 1420.4 $\pm$ 4.7 & 29 \\
\bottomrule
\end{tabular}
}
\caption{Held-out-topic roster (development profiles, new topics; \numSeedsT{} seeds) and development roster (\numSeedsD{} seeds, used for design, descriptive): same layout as Table~\ref{tab:main}. The floor $G_T$, already subtracted from every entry: 659 (held-out topics), 568 (development).}
\label{tab:topic}
\end{table}

%% file: generated/tab_topic_gains.tex
\begin{table}[t]
\centering
\small\setlength{\tabcolsep}{4pt}
\resizebox{\linewidth}{!}{%
\begin{tabular}{lrr}
\toprule
compared with & held-out topics & development \\
\midrule
RUCB (pooled, non-contextual) & +21.7\% [+20.3\%, +23.3\%] & +22.7\% [+20.7\%, +24.6\%] \\
Pooled & +23.9\% [+21.1\%, +26.9\%] & +29.9\% [+22.7\%, +38.8\%] \\
CoLSTIM & +33.4\% [+32.2\%, +34.5\%] & +29.7\% [+28.1\%, +31.3\%] \\
COLDB & +17.9\% [+16.3\%, +19.5\%] & +18.3\% [+16.3\%, +20.4\%] \\
COLDB-LR & +18.9\% [+17.5\%, +20.5\%] & +19.4\% [+17.5\%, +21.2\%] \\
Meta-TS & +20.3\% [+18.8\%, +22.0\%] & +20.8\% [+19.0\%, +22.6\%] \\
HierTS & +18.8\% [+17.5\%, +20.4\%] & +19.1\% [+17.5\%, +20.7\%] \\
AdaTS & +20.5\% [+18.2\%, +23.2\%] & +19.7\% [+18.1\%, +21.4\%] \\
\textit{type-library reference} & -1.1\% [-2.7\%, +0.6\%] & -2.0\% [-4.0\%, +0.0\%] \\
\bottomrule
\end{tabular}
}
\caption{Held-out-topic roster and development roster: same layout as Table~\ref{tab:main-gains}.}
\label{tab:topic-gains}
\end{table}

%% file: generated/tab_phases.tex
\begin{table}[t]
\centering
\small\setlength{\tabcolsep}{4pt}
\begin{tabular}{lrrrrrr}
\toprule
method & \multicolumn{2}{c}{held-out profiles} & \multicolumn{2}{c}{Composite} & \multicolumn{2}{c}{Quirky} \\
 & first 25\% & last 25\% & first 25\% & last 25\% & first 25\% & last 25\% \\
\midrule
no transfer & 4.61 & 4.65 & 4.74 & 4.75 & 4.77 & 4.80 \\
\method{} & 3.90 & 3.19 & 4.08 & 3.45 & 4.01 & 3.37 \\
\textit{type-library reference} & 5.10 & 5.16 & 4.42 & 4.42 & 4.38 & 4.42 \\
\bottomrule
\end{tabular}
\caption{Excess regret per user in the first and last quarter of the arrivals (mean over seeds).}
\label{tab:phases}
\end{table}

%% file: generated/tab_sim.tex
\begin{table}[H]
\centering
\footnotesize\setlength{\tabcolsep}{3pt}
\resizebox{\linewidth}{!}{%
\begin{tabular}{lrrrrrrrrrrr}
\toprule
population & no transfer & RUCB & CoLSTIM & COLDB & COLDB-LR & Pooled & Meta-TS & HierTS & AdaTS & \method{} & oracle \\
\midrule
\multicolumn{12}{l}{\emph{$\beta=4$}} \\
\quad 2 types, $60^\circ$ & 936 $\pm$ 6 & 2066 $\pm$ 22 & 911 $\pm$ 7 & 288 $\pm$ 9 & 260 $\pm$ 10 & 298 $\pm$ 10 & 332 $\pm$ 9 & 322 $\pm$ 9 & 325 $\pm$ 9 & 178 $\pm$ 3 & 105 $\pm$ 3 \\
\quad 2 types, $90^\circ$ & 930 $\pm$ 5 & 2045 $\pm$ 25 & 905 $\pm$ 6 & 624 $\pm$ 19 & 308 $\pm$ 7 & 644 $\pm$ 20 & 675 $\pm$ 18 & 660 $\pm$ 19 & 664 $\pm$ 18 & 219 $\pm$ 2 & 156 $\pm$ 4 \\
\quad 4 types, $60^\circ$ & 934 $\pm$ 4 & 2081 $\pm$ 18 & 907 $\pm$ 5 & 440 $\pm$ 11 & 472 $\pm$ 10 & 454 $\pm$ 11 & 487 $\pm$ 11 & 476 $\pm$ 11 & 477 $\pm$ 11 & 297 $\pm$ 5 & 208 $\pm$ 4 \\
\quad 4 types, $90^\circ$ & 932 $\pm$ 4 & 2061 $\pm$ 16 & 905 $\pm$ 5 & 1051 $\pm$ 18 & 600 $\pm$ 10 & 1075 $\pm$ 19 & 1101 $\pm$ 19 & 1083 $\pm$ 18 & 1087 $\pm$ 18 & 401 $\pm$ 4 & 335 $\pm$ 5 \\
\quad 8 types, $60^\circ$ & 926 $\pm$ 4 & 2068 $\pm$ 16 & 903 $\pm$ 5 & 537 $\pm$ 7 & 669 $\pm$ 9 & 551 $\pm$ 8 & 586 $\pm$ 8 & 574 $\pm$ 8 & 576 $\pm$ 8 & 416 $\pm$ 4 & 304 $\pm$ 4 \\
\quad 8 types, $90^\circ$ & 927 $\pm$ 3 & 2075 $\pm$ 8 & 905 $\pm$ 4 & 1377 $\pm$ 13 & 959 $\pm$ 12 & 1416 $\pm$ 12 & 1416 $\pm$ 11 & 1405 $\pm$ 12 & 1396 $\pm$ 11 & 618 $\pm$ 5 & 526 $\pm$ 5 \\
\quad Gaussian cloud & 850 $\pm$ 4 & 1770 $\pm$ 14 & 820 $\pm$ 5 & 868 $\pm$ 10 & 877 $\pm$ 8 & 890 $\pm$ 11 & 934 $\pm$ 10 & 908 $\pm$ 11 & 919 $\pm$ 10 & 611 $\pm$ 6 & 492 $\pm$ 5 \\
\quad low-rank cloud & 841 $\pm$ 6 & 1763 $\pm$ 20 & 813 $\pm$ 7 & 1157 $\pm$ 24 & 641 $\pm$ 14 & 1176 $\pm$ 26 & 1221 $\pm$ 23 & 1200 $\pm$ 25 & 1210 $\pm$ 23 & 401 $\pm$ 7 & 403 $\pm$ 7 \\
\quad null population & 865 $\pm$ 4 & 1823 $\pm$ 7 & 835 $\pm$ 4 & 1792 $\pm$ 9 & 1283 $\pm$ 7 & 1836 $\pm$ 11 & 1671 $\pm$ 7 & 1702 $\pm$ 7 & 1629 $\pm$ 6 & 867 $\pm$ 3 & 859 $\pm$ 5 \\
\midrule
\multicolumn{12}{l}{\emph{$\beta=12$}} \\
\quad 2 types, $60^\circ$ & 664 $\pm$ 5 & 2068 $\pm$ 23 & 655 $\pm$ 5 & 282 $\pm$ 9 & 124 $\pm$ 5 & 293 $\pm$ 9 & 314 $\pm$ 7 & 292 $\pm$ 8 & 308 $\pm$ 7 & 102 $\pm$ 1 & 40 $\pm$ 1 \\
\quad 2 types, $90^\circ$ & 665 $\pm$ 4 & 2049 $\pm$ 25 & 654 $\pm$ 3 & 625 $\pm$ 20 & 154 $\pm$ 5 & 644 $\pm$ 20 & 635 $\pm$ 14 & 616 $\pm$ 16 & 660 $\pm$ 42 & 111 $\pm$ 1 & 62 $\pm$ 2 \\
\quad 4 types, $60^\circ$ & 662 $\pm$ 4 & 2081 $\pm$ 17 & 652 $\pm$ 4 & 435 $\pm$ 11 & 203 $\pm$ 4 & 448 $\pm$ 11 & 461 $\pm$ 10 & 443 $\pm$ 10 & 455 $\pm$ 10 & 166 $\pm$ 3 & 90 $\pm$ 2 \\
\quad 4 types, $90^\circ$ & 663 $\pm$ 3 & 2061 $\pm$ 17 & 654 $\pm$ 3 & 1044 $\pm$ 18 & 276 $\pm$ 3 & 1072 $\pm$ 19 & 998 $\pm$ 13 & 1009 $\pm$ 16 & 1018 $\pm$ 32 & 215 $\pm$ 2 & 142 $\pm$ 2 \\
\quad 8 types, $60^\circ$ & 663 $\pm$ 3 & 2067 $\pm$ 16 & 658 $\pm$ 4 & 527 $\pm$ 7 & 308 $\pm$ 4 & 544 $\pm$ 7 & 556 $\pm$ 7 & 532 $\pm$ 7 & 547 $\pm$ 7 & 237 $\pm$ 3 & 139 $\pm$ 2 \\
\quad 8 types, $90^\circ$ & 659 $\pm$ 3 & 2073 $\pm$ 8 & 655 $\pm$ 3 & 1367 $\pm$ 12 & 434 $\pm$ 5 & 1406 $\pm$ 13 & 1222 $\pm$ 7 & 1281 $\pm$ 10 & 1213 $\pm$ 7 & 337 $\pm$ 2 & 244 $\pm$ 3 \\
\quad Gaussian cloud & 585 $\pm$ 3 & 1770 $\pm$ 14 & 575 $\pm$ 3 & 855 $\pm$ 10 & 613 $\pm$ 6 & 878 $\pm$ 10 & 848 $\pm$ 8 & 847 $\pm$ 9 & 839 $\pm$ 8 & 448 $\pm$ 5 & 334 $\pm$ 4 \\
\quad low-rank cloud & 586 $\pm$ 4 & 1762 $\pm$ 21 & 574 $\pm$ 4 & 1142 $\pm$ 24 & 382 $\pm$ 6 & 1167 $\pm$ 25 & 1061 $\pm$ 14 & 1111 $\pm$ 18 & 1045 $\pm$ 14 & 249 $\pm$ 5 & 203 $\pm$ 4 \\
\quad null population & 598 $\pm$ 3 & 1825 $\pm$ 7 & 587 $\pm$ 3 & 1783 $\pm$ 11 & 890 $\pm$ 5 & 1837 $\pm$ 10 & 1301 $\pm$ 6 & 1426 $\pm$ 6 & 1279 $\pm$ 5 & 601 $\pm$ 3 & 598 $\pm$ 3 \\
\bottomrule
\end{tabular}
}
\caption{Simulator, $b=20$, $u=300$: excess regret (mean $\pm$ s.e.\ over seeds) by population and policy. COLDB-LR is our likelihood-ratio variant of COLDB; the population oracle starts each user from the true $\mathcal{P}$.}
\label{tab:sim}
\end{table}

%% file: generated/tab_tuning.tex
\begin{table}[t]
\centering
\small\setlength{\tabcolsep}{3pt}
\textbf{(a) Search grids}\par\smallskip
\begin{tabular}{@{}p{2.1cm}p{5.35cm}p{5.35cm}@{}}
\toprule
method & tutoring & simulator \\
\midrule
CoLSTIM & $c\in\{0,\allowbreak 0.3,\allowbreak 1,\allowbreak 3\}$\newline $\ell\in\{0.03,\allowbreak 0.1,\allowbreak 0.3,\allowbreak 1\}$\newline $\gamma\in\{0.1,\allowbreak 0.3,\allowbreak 1,\allowbreak 3\}$\newline 64 configurations & $c\in\{0,\allowbreak 0.3,\allowbreak 1,\allowbreak 3\}$\newline $\ell\in\{0.3,\allowbreak 1,\allowbreak 3,\allowbreak 10\}$\newline $\gamma\in\{0.01,\allowbreak 0.03,\allowbreak 0.1,\allowbreak 0.3,\allowbreak 1,\allowbreak 3\}$\newline 96 configurations \\
COLDB & $\alpha\in\{0.001,\allowbreak 0.003,\allowbreak 0.01,\allowbreak 0.03,\allowbreak 0.1,\allowbreak 0.3,\allowbreak 1,\allowbreak 3\}$\newline $c\in\{0,\allowbreak 0.03,\allowbreak 0.1,\allowbreak 0.3,\allowbreak 1\}$\newline 40 configurations & $\alpha\in\{0.03,\allowbreak 0.1,\allowbreak 0.3,\allowbreak 1,\allowbreak 3\}$\newline $c\in\{0,\allowbreak 0.1,\allowbreak 0.3,\allowbreak 1\}$\newline 20 configurations \\
COLDB-LR & $\mathrm{tol}\in\{0.25,\allowbreak 0.5,\allowbreak 1,\allowbreak 2,\allowbreak 4\}$\newline $c\in\{0,\allowbreak 0.1,\allowbreak 0.3,\allowbreak 1\}$\newline 20 configurations & $\mathrm{tol}\in\{0.25,\allowbreak 0.5,\allowbreak 1,\allowbreak 2,\allowbreak 4\}$\newline $c\in\{0,\allowbreak 0.1,\allowbreak 0.3,\allowbreak 1\}$\newline 20 configurations \\
RUCB & $\alpha\in\{0.51,\allowbreak 1,\allowbreak 2,\allowbreak 4\}$\newline 4 configurations & $\alpha\in\{0.51,\allowbreak 1,\allowbreak 2,\allowbreak 4\}$\newline 4 configurations \\
Meta-TS & $s_0\in\{0.0003,\allowbreak 0.001,\allowbreak 0.003,\allowbreak 0.01,\allowbreak 0.03,\allowbreak 0.1,\allowbreak 0.3,\allowbreak 1,\allowbreak 3\}$\newline $s_q\in\{0.001,\allowbreak 0.01,\allowbreak 0.1,\allowbreak 1,\allowbreak 10,\allowbreak 100,\allowbreak 1000\}$\newline 63 configurations & $s_0\in\{0.0003,\allowbreak 0.001,\allowbreak 0.003,\allowbreak 0.01,\allowbreak 0.03,\allowbreak 0.1,\allowbreak 0.3,\allowbreak 1,\allowbreak 3\}$\newline $s_q\in\{0.001,\allowbreak 0.01,\allowbreak 0.1,\allowbreak 1,\allowbreak 10,\allowbreak 100,\allowbreak 1000\}$\newline 63 configurations \\
HierTS & $s_0\in\{0.0003,\allowbreak 0.001,\allowbreak 0.003,\allowbreak 0.01,\allowbreak 0.03,\allowbreak 0.1,\allowbreak 0.3,\allowbreak 1,\allowbreak 3\}$\newline $s_q\in\{0.1,\allowbreak 1,\allowbreak 10,\allowbreak 100,\allowbreak 1000\}$\newline 45 configurations & $s_0\in\{0.0003,\allowbreak 0.001,\allowbreak 0.003,\allowbreak 0.01,\allowbreak 0.03,\allowbreak 0.1,\allowbreak 0.3,\allowbreak 1,\allowbreak 3\}$\newline $s_q\in\{0.001,\allowbreak 0.01,\allowbreak 0.1,\allowbreak 1,\allowbreak 10,\allowbreak 100,\allowbreak 1000\}$\newline 63 configurations \\
AdaTS & $s_0\in\{0.0003,\allowbreak 0.001,\allowbreak 0.003,\allowbreak 0.01,\allowbreak 0.03,\allowbreak 0.1,\allowbreak 0.3,\allowbreak 1,\allowbreak 3\}$\newline $s_q\in\{0.1,\allowbreak 1,\allowbreak 10,\allowbreak 100,\allowbreak 1000\}$\newline 45 configurations & $s_0\in\{0.0003,\allowbreak 0.001,\allowbreak 0.003,\allowbreak 0.01,\allowbreak 0.03,\allowbreak 0.1,\allowbreak 0.3,\allowbreak 1,\allowbreak 3\}$\newline $s_q\in\{0.001,\allowbreak 0.01,\allowbreak 0.1,\allowbreak 1,\allowbreak 10,\allowbreak 100,\allowbreak 1000\}$\newline 63 configurations \\
Latent-bandit UCB (\method{} ablation) & $c_{\mathrm{UCB}}\in\{0.25,\allowbreak 0.5,\allowbreak 1,\allowbreak 2,\allowbreak 3,\allowbreak 5\}$\newline 6 configurations & -- \\
\bottomrule
\end{tabular}
\par\medskip\textbf{(b) Selected settings}\par\smallskip
\begin{tabular}{@{}p{2.1cm}p{4.25cm}rp{4.25cm}r@{}}
\toprule
method & tutoring setting & ratio & simulator setting & ratio \\
\midrule
CoLSTIM & $c=1$, $\ell=0.1$, $\gamma=0.3$ & 0.963 & $c=0.3$, $\ell=3$, $\gamma=0.01$ & 0.978 \\
COLDB & $\alpha=0.03$, $c=0.03$ & 0.828 & $\alpha=0.1$, $c=0$ & 1.071 \\
COLDB-LR & $\mathrm{tol}=2$, $c=0.3$ & 0.846 & $\mathrm{tol}=0.5$, $c=0.3$ & 0.577 \\
RUCB & $\alpha=0.51$ & 0.896 & $\alpha=1$ & 2.660 \\
Meta-TS & $s_0=0.001$, $s_q=0.1$ & 0.853 & $s_0=0.03$, $s_q=1$ & 1.073 \\
HierTS & $s_0=0.001$, $s_q=1$ & 0.840 & $s_0=0.01$, $s_q=0.1$ & 1.068 \\
AdaTS & $s_0=0.0003$, $s_q=10$ & 0.840 & $s_0=0.03$, $s_q=1$ & 1.057 \\
Latent-bandit UCB (\method{} ablation) & $c_{\mathrm{UCB}}=0.5$ & 0.773 & -- & -- \\
\bottomrule
\end{tabular}
\caption{Search grids and selected settings. Every configuration uses $16$ tuning seeds, separate from benchmark evaluation: the development roster ($b=24$) for tutoring and the $16$ structured $b=20$ settings for the simulator. The ratio is excess regret divided by that of no transfer, averaged over settings (lower is better). Here $c$ scales exploration, $\ell$ multiplies the base ridge $d/B_\theta^2$, and $\gamma$ scales CoLSTIM's Gumbel perturbations (clipped to $[-3,3]$ before scaling). For COLDB, $\alpha$ scales the distance threshold; for RUCB, it controls the confidence width. $\mathrm{tol}$ is COLDB-LR's likelihood tolerance. The variance multipliers $s_0$ and $s_q$ scale $v_0=B_\theta^2/d$ for user variation and the population-mean prior, respectively. $c_{\mathrm{UCB}}$ is the log-weight margin used to select plausible components in the latent-bandit UCB ablation. Methods use the supplied features without appending additional persona indicators.}
\label{tab:tuning}
\end{table}

%% file: generated/tab_tuning_oracle.tex
\begin{table}[t]
\centering
\small\setlength{\tabcolsep}{3pt}
\resizebox{\linewidth}{!}{%
\begin{tabular}{lrrlrrlrrl}
\toprule
method & \multicolumn{3}{c}{held-out profiles} & \multicolumn{3}{c}{Composite} & \multicolumn{3}{c}{Quirky} \\
 & dev.\ choice & best on roster & its configuration & dev.\ choice & best on roster & its configuration & dev.\ choice & best on roster & its configuration \\
\midrule
CoLSTIM & 33 & 34 & \texttt{c1:l0.1:g0.1} & 28 & 28 & \texttt{c1:l0.3:g0.3} & 28 & 29 & \texttt{c1:l0.1:g0.1} \\
COLDB & 22 & 32 & \texttt{a0.01:c0.03} & 38 & 39 & \texttt{a0.03:c0} & 38 & 40 & \texttt{a0.03:c0.1} \\
COLDB-LR & 22 & 34 & \texttt{t0.25:c0.3} & 34 & 34 & \texttt{t2:c0.1} & 36 & 36 & \texttt{t2:c0.3} \\
RUCB & 17 & 17 & \texttt{a0.51} & 32 & 32 & \texttt{a0.51} & 34 & 34 & \texttt{a0.51} \\
Meta-TS & 23 & 23 & \texttt{s0=3:sq=10} & 34 & 36 & \texttt{s0=0.0003:sq=1} & 35 & 38 & \texttt{s0=0.003:sq=1} \\
HierTS & 23 & 24 & \texttt{s0=0.01:sq=1} & 37 & 37 & \texttt{s0=0.001:sq=1} & 38 & 38 & \texttt{s0=0.003:sq=1000} \\
AdaTS & 22 & 24 & \texttt{s0=0.01:sq=1} & 37 & 37 & \texttt{s0=0.003:sq=100} & 36 & 38 & \texttt{s0=0.0003:sq=1000} \\
\midrule
\method{} & 48 & &  & 44 & &  & 46 & &  \\
\bottomrule
\end{tabular}
}
\caption{Benchmark score of each comparison method with the configuration chosen on the development roster (dev.\ choice, as in Table~\ref{tab:main}) and with the configuration of its grid (Table~\ref{tab:tuning}) that has the lowest excess regret on the evaluation roster itself (best on roster, chosen on the evaluation seeds and therefore optimistic), with that configuration. The last row repeats \method{}'s score.}
\label{tab:tuning-oracle}
\end{table}

%% file: generated/tab_ts_pair.tex
\begin{table}[t]
\centering
\small
\begin{tabular}{lrrr}
\toprule
method & held-out profiles & Composite & Quirky \\
\midrule
Meta-TS & -1.3 [-2.0, -0.6] & +0.8 [-1.9, +3.5] & -0.6 [-3.6, +2.5] \\
HierTS & -0.1 [-0.7, +0.5] & +1.5 [+0.4, +2.5] & -1.7 [-3.7, +0.2] \\
AdaTS & +0.8 [-1.6, +3.2] & +1.3 [-1.2, +3.7] & -2.1 [-6.3, +2.2] \\
\bottomrule
\end{tabular}
\caption{Prior-learning Thompson samplers with a double-Thompson-sampling challenger (the second persona maximizes an independent second posterior sample, prior variances re-tuned on the development roster) against the reported rule (the two highest utilities of one sample): paired change in excess regret as a share of the reported version's excess regret, with a 95\% $t$ interval over seeds. Negative values favor the double-sample rule.}
\label{tab:ts-pair}
\end{table}

%% file: generated/tab_references.tex
\begin{table}[!htbp]
\centering
\small\setlength{\tabcolsep}{4pt}
\resizebox{\linewidth}{!}{%
\begin{tabular}{lrrrrr}
\toprule
reference & held-out profiles & Composite & Quirky & held-out topics & development \\
\midrule
\textit{best fixed pair per user} & 76 & 66 & 68 & 74 & 73 \\
\textit{known-profile pair} & 75 & -- & -- & 73 & 72 \\
\textit{type-library reference} & 23 & 32 & 33 & 54 & 52 \\
\bottomrule
\end{tabular}
}
\caption{Diagnostic references, by roster: benchmark score, the share of random pairs' excess regret removed (higher is better). The best fixed pair per user shows, for the whole session, the pair with the lowest regret on that user's questions in hindsight. The known-profile pair shows the best fixed pair of the user's profile, chosen on the development roster for the development profiles and cross-fitted for the held-out profiles; it is not defined on Composite and Quirky. The type-library reference knows each development profile's average choice probabilities and utilities and infers the current user's profile from their choices.}
\label{tab:references}
\end{table}

%% file: sections/appendix_ablations.tex
\section{Ablations and Component Studies}
\label{app:ablations}

We change one component of \method{} at a time on the development roster ($b=24$), keeping every other parameter of the reported configuration. All policies see the same users, questions and clicks for a given seed. Regret values are excess strong regret $R_T-G_T$ on the common evaluation scale. These studies are descriptive: the development roster was also used for design.

\subsection{One-component changes}

\input{generated/tab_ablations}

Table~\ref{tab:ablations} reports how much the excess regret of \method{} rises when a component is changed. Its first row removes every component that transfers information across users, leaving the no-transfer learner (\numAblNoTransfer{}). Removing only the population, so that the same learner, with the recalibrated scores, restarts for every user, costs the most (\numAblNoPop{}). One tight type instead of the mixture costs \numAblOneComp{}, while one Gaussian with learned covariance is within noise of the mixture (\numAblOneGauss{}, 95\% interval \numAblOneGaussLo{} to \numAblOneGaussHi{}). Without recalibration, excess regret rises by \numAblNoRecal{} (95\% interval \numAblNoRecalLo{} to \numAblNoRecalHi{}), and a smaller fallback concentration raises it by \numAblAlphaFive{} (\numAblAlphaFiveLo{} to \numAblAlphaFiveHi{}). Latent-bandit Thompson sampling and UCB over the same components~\citep{hong2020latent} cost \numAblLts{} and \numAblLucb{}, because a sampled or optimistic component's favorite pair is paid in strong regret whenever that component is wrong. The greedy pair rule (\numAblGreedy{}, 95\% interval \numAblGreedyLo{} to \numAblGreedyHi{}), the moment M-step (\numAblMoment{}) and removing the evidence gate (\numAblNoGate{}) are each within noise of the reported configuration on this roster.

\subsection{Component effects on held-out rosters}

\input{generated/tab_ablations_heldout}

Table~\ref{tab:ablations-heldout} repeats the main component changes on the three evaluation rosters of Table~\ref{tab:main}, paired with the runs of that table. These rosters were not used for any design choice. These comparisons test whether the component effects observed during development persist for new students. The learned population carries most of what \method{} adds on every roster: without it, excess regret rises by \numAblHNoPopUShareAbs{}, \numAblHNoPopCShareAbs{}, and \numAblHNoPopIShareAbs{}. One tight type raises it by \numAblHOneCompUShareAbs{}, \numAblHOneCompCShareAbs{}, and \numAblHOneCompIShareAbs{}. One Gaussian with learned covariance raises it by \numAblHOneGaussUShareAbs{} on the held-out profiles and lowers it by \numAblHOneGaussCShareAbs{} on Composite and \numAblHOneGaussIShareAbs{} on Quirky, each 95\% interval excluding zero. Without recalibration, excess regret rises by \numAblHNoRecalUShareAbs{}, \numAblHNoRecalCShareAbs{}, and \numAblHNoRecalIShareAbs{}. Greedy pairs raise it by \numAblHGreedyUShareAbs{} on the held-out profiles, lower it by \numAblHGreedyCShareAbs{} on Composite, and change it within noise on Quirky. Removing the gate lowers excess regret by \numAblHNoGateCShareAbs{} on Composite and changes it within noise on the other two rosters. Learning a population and recalibrating scores help on all three rosters, while the benefits of the mixture, lookahead, and gate depend on the roster.

\subsection{Sensitivity to the settings of \method{}}

\input{generated/tab_sensitivity}

The settings of \method{} were fixed on the development roster. Table~\ref{tab:sensitivity} changes one of them at a time on the three evaluation rosters, halving and doubling the gate threshold $z$, the gate's minimum number of completed sessions $n_{\min}$, the maximum number of components $M_{\max}$, the question-bank capacity, and the recalibration's ridge and refit interval, and setting $\alpha$ to $5$ and $50$. These runs are descriptive and were not used to choose any setting. Across the \numSensCells{} changed settings and rosters, the score moves by \numSensWorst{} to \numSensBest{} points, and in \numSensBelowClosest{} of them \method{} keeps lower excess regret than the closest comparison method on that roster (significantly in \numSensSigBelowClosest{}).

\subsection{Evidence gate}

\input{generated/tab_controls}

\input{generated/tab_ablation_gate}

Table~\ref{tab:ablation-gate} evaluates the same policy with and without its gate on the two negative controls, together with the learner without a population. The gate lets the population learner choose pairs only while, on completed users, it predicts choices better than the no-transfer learner by more than $z=0.5$ standard errors (Section~\ref{sec:algorithm}). On the simulator's null population, where users share no structure, the gate keeps \method{} at the level of no transfer (\numGateSimNullGatedShare{}, against \numGateSimNullUngatedShare{} without it): without the gate, the learned population pulls new users toward patterns that do not exist. On the scrambled tutoring roster, the gated and ungated policies both stay within one percent of no transfer (\numGateScrambledGatedShare{} and \numGateScrambledUngatedShare{}). We set the threshold on the development roster, before running either control.

\subsection{Lookahead with a known population}

\input{generated/tab_ablation_library}

To separate learning the population from choosing informative pairs, this diagnostic gives the pair rule the development profiles' choice probabilities and utilities. It is not an admissible learning policy. With the profiles known, the one-step lookahead and the greedy rule are close (\numLibLookahead{} and \numLibGreedy{}), while latent-bandit Thompson sampling (\numLibLts{}) and UCB (\numLibLucb{}) are far above greedy, for the reason given above.

\subsection{Implementation details}
\label{app:coduel-impl}

Algorithm~\ref{alg:population-transfer} summarizes \method{} (Section~\ref{sec:algorithm}).

\begin{algorithm}[H]
\caption{\method{}}
\label{alg:population-transfer}
\small
\begin{algorithmic}[1]
\REQUIRE Parameters $d,K,B_\theta,\beta,b,u$ and candidate features $\phi(p,q)$; constants $\alpha,M_{\max},z,n_{\min}$.
\STATE $v_0\gets B_\theta^2/d$; $\Pi_0\gets\mathcal{N}(0,v_0 I_d)$; gate $\gets$ closed; evidence list $L\gets\varnothing$; $A\gets I_K$.
\STATE Initialize an empty comparison log and question bank $\mathcal{Q}_0\gets\varnothing$.
\FOR{user $i=1,\ldots,u$}
    \STATE Initialize the no-transfer learner's posterior to $\mathcal{N}(0,v_0 I_d)$ and the population learner's belief to $\Pi_{i-1}$; $\ell\gets0$.
    \FOR{round $t=(i-1)b+1,\ldots,ib$}
        \STATE Observe $q_t$ and all $\phi(p,q_t)$; retain the original features in user $i$'s log. If $\phi$ contains a rater score, replace $g(p,q_t)$ by the centered and rescaled $(A\,g(q_t))_p$.
        \STATE Append $q_t$ to $\mathcal{Q}_{t-1}$ and retain the $16$ most recent questions to form $\mathcal{Q}_t$.
        \STATE $(p^{\mathrm{nt}}_{t,1},p^{\mathrm{nt}}_{t,2})\gets$ the two highest-ranked personas under the no-transfer learner's posterior mean.
        \IF{gate is open}
            \STATE $(p_{t,1},p_{t,2})\gets\arg\min J_t$ using~(\ref{eq:transfer-gate}) (greedy in the last round).
        \ELSE
            \STATE $(p_{t,1},p_{t,2})\gets(p^{\mathrm{nt}}_{t,1},p^{\mathrm{nt}}_{t,2})$.
        \ENDIF
        \STATE Record both learners' predicted probabilities that $p_{t,1}$ is chosen; display the pair; observe $y_t$.
        \STATE Store the displayed pair and $y_t$ in user $i$'s log for $q_t$.
        \STATE $\ell\gets\ell+\log\frac{\Pr_{\mathrm{pop}}(y_t)}{\Pr_{\mathrm{nt}}(y_t)}$; update both posteriors with the displayed comparison and~(\ref{eq:btl}).
    \ENDFOR
    \IF{$\phi$ contains a rater score, $i\geq20$, and $i$ is a multiple of $10$}
        \STATE Refit $A$ on the completed sessions' choices; rebuild both learners' features and posteriors and the question bank's features from the log, keeping $L$ unchanged.
    \ENDIF
    \STATE Append $\ell$ to $L$; if $|L|\geq n_{\min}$, open the gate iff $\mathrm{mean}(L)>z\cdot\mathrm{s.e.}(L)$, else close it.
    \STATE Run EM, including new components, on the completed sessions; construct $\Pi_i$ using~(\ref{eq:transfer-prior}).
\ENDFOR
\end{algorithmic}
\end{algorithm}

\textbf{Posterior updates and predictions.}
The no-transfer learner and each component of the population learner use a Laplace approximation: a Gaussian centered at the posterior mode, with covariance equal to the inverse Hessian of the negative log posterior there, recomputed after each choice from all of the current session's comparisons. Pair rules break exact ties uniformly at random. For a candidate pair $(p,p')$ on $q_t$, component $k$ of the population learner's belief predicts the probability that $p$ is chosen as
\begin{equation*}
\rho_{k,t}(p,p')\approx\mathbb{E}_{w\sim\mathcal{N}(m_{k,t},C_{k,t})}\!\left[\sigma\!\left(\beta[\phi(p,q_t)-\phi(p',q_t)]^\top w\right)\right].
\end{equation*}
where $\sigma$ is the logistic function of~(\ref{eq:btl}); we evaluate the expectation with a probit approximation.

The EM procedure holds the isotropic component's mean and covariance fixed. Its M-step pools the weighted comparisons instead of averaging the users' posterior means, because a narrow component prior can keep users' posterior means near its current mean and so limit updates based on their average (the moment M-step of Table~\ref{tab:ablations} averages them). We run three EM-style iterations after each completed session. The first pass starts from a single learned Gaussian with the mean and covariance of the first session's Laplace posterior under the isotropic prior. Each pooled ridge logistic fit uses prior precision $I_d/v_0$. Learned component covariances receive a $10^{-3}I_d$ ridge for numerical stability. We remove learned components whose total responsibility across users is below one, retaining the component with the largest total if all fall below this threshold. We normalize the retained totals to obtain $\pi_k$.

After each pass, we select the session with the largest positive difference between its approximate log marginal likelihood under the isotropic component and the maximum over learned components. If such a session exists, we initialize a new component from its Laplace posterior under the isotropic prior and run three more EM-style iterations. We repeat this process, with at most three such additions per session and at most $16$ learned components. While no learned component exists, the lookahead reduces to the greedy rule.

The recalibration fits a logistic model of each completed comparison: the log-odds that $p_{t,1}$ is chosen equal $(e_{p_{t,1}}-e_{p_{t,2}})^\top(\xi+\xi_{i_t})+(A\,g(q_t))_{p_{t,1}}-(A\,g(q_t))_{p_{t,2}}$, where $\xi\in\mathbb{R}^K$ is the persona offset shared by all users and $\xi_i$ the offset of user $i$. Besides the penalty on $A'$ (Section~\ref{sec:algorithm}), the ridge penalties are $10^{-3}$ on $\xi$ and $a$ and $1$ on each $\xi_i$, and each refit starts from the previous solution. The recalibrated coordinate is centered over the six personas of each question before rescaling. When $A$ changes, we recompute the stored features of all completed sessions, the current session's posteriors, the lookahead's question bank, and the no-transfer learner's records, and keep the gate's evidence list.

In the tutoring benchmark, $B_\theta$ is a prior scale rather than a known bound: it sets the initial prior variance $v_0=B_\theta^2/d$ of Algorithm~\ref{alg:population-transfer}. We set it to the median Euclidean norm of per-user linear fits of the choice logits on the feature contrasts on the development roster ($B_\theta=1.84$; $2.70$ for the second judge, Appendix~\ref{app:second-judge}).

%% file: generated/tab_ablations.tex
\begin{table}[t]
\centering
\small
\begin{tabular}{lr}
\toprule
variant & increase in excess regret [95\% CI] \\
\midrule
\method{} (excess regret) & 970.8 $\pm$ 4.7 \\
\midrule
no transfer (base learner only) & +449.7 [+434.7, +464.6] \\
greedy pair rule instead of lookahead & +5.2 [-7.9, +18.2] \\
latent-bandit Thompson sampling instead of lookahead & +213.0 [+196.8, +229.1] \\
latent-bandit UCB instead of lookahead & +128.5 [+109.4, +147.6] \\
no evidence gate & +8.1 [-4.9, +21.1] \\
fallback concentration $\alpha=5$ & +22.1 [+3.8, +40.3] \\
moment M-step instead of pooled fit & +3.7 [-14.3, +21.7] \\
one tight type instead of the mixture (no spread between users) & +157.7 [+144.9, +170.5] \\
one Gaussian with learned covariance instead of the mixture & -2.5 [-14.3, +9.3] \\
no population (recalibration kept) & +386.2 [+374.9, +397.4] \\
no recalibration ($A=I_K$) & +41.2 [+30.2, +52.2] \\
\bottomrule
\end{tabular}
\caption{One-component changes of \method{} on the development roster (\numAblSeeds{} seeds, $b=24$). First row: excess regret of \method{} (mean $\pm$ s.e.). Other rows: paired increase in excess regret when the component is changed, with a 95\% $t$ interval over seeds; positive values mean the component of \method{} lowers regret. Every other parameter is unchanged.}
\label{tab:ablations}
\end{table}

%% file: generated/tab_ablations_heldout.tex
\begin{table}[t]
\centering
\small\setlength{\tabcolsep}{4pt}
\resizebox{\linewidth}{!}{%
\begin{tabular}{lrrr}
\toprule
variant & held-out profiles & Composite & Quirky \\
\midrule
greedy pair rule instead of lookahead & +46.6 [+33.7, +59.5] & -18.0 [-33.4, -2.7] & -1.7 [-16.6, +13.2] \\
no evidence gate & +3.1 [-9.8, +16.0] & -19.1 [-33.4, -4.8] & -7.8 [-24.7, +9.1] \\
one tight type instead of the mixture (no spread between users) & +359.0 [+344.1, +374.0] & +120.6 [+102.2, +139.0] & +147.2 [+127.3, +167.0] \\
one Gaussian with learned covariance instead of the mixture & +84.4 [+71.5, +97.2] & -28.1 [-46.8, -9.3] & -29.7 [-43.7, -15.7] \\
no population (recalibration kept) & +370.4 [+359.1, +381.7] & +368.5 [+346.2, +390.8] & +384.1 [+367.9, +400.3] \\
no recalibration ($A=I_K$) & +33.7 [+21.6, +45.9] & +29.7 [+17.2, +42.2] & +50.7 [+34.9, +66.5] \\
\bottomrule
\end{tabular}
}
\caption{One-component changes of \method{} on the three main rosters (\numSeedsU{} seeds on the held-out profiles, \numAblHSeedsC{} on Composite and Quirky): paired increase in excess regret when the component is changed, with a 95\% $t$ interval over seeds (roster fixed); positive values mean the component of \method{} lowers regret; negative values favor the variant.}
\label{tab:ablations-heldout}
\end{table}

%% file: generated/tab_sensitivity.tex
\begin{table}[t]
\centering
\small\setlength{\tabcolsep}{4pt}
\resizebox{\linewidth}{!}{%
\begin{tabular}{lrrr}
\toprule
setting & held-out profiles & Composite & Quirky \\
\midrule
gate threshold $z=0.25$ (paper $0.5$) & -0.1 [-0.6, +0.5] & +0.6 [-0.1, +1.4] & +0.0 [-0.8, +0.9] \\
gate threshold $z=1$ & -0.3 [-0.9, +0.2] & +0.5 [-0.2, +1.3] & +0.2 [-0.4, +0.7] \\
gate minimum $n_{\min}=4$ (paper $8$) & +0.1 [-0.5, +0.6] & +0.6 [-0.1, +1.2] & +0.3 [-0.3, +0.9] \\
gate minimum $n_{\min}=16$ & -0.7 [-1.2, -0.1] & +0.5 [-0.2, +1.1] & +0.1 [-0.5, +0.7] \\
components $M_{\max}=8$ (paper $16$) & -0.4 [-0.9, +0.1] & +0.6 [-0.1, +1.3] & +0.0 [-0.6, +0.7] \\
components $M_{\max}=32$ & -0.3 [-0.9, +0.3] & +0.6 [+0.0, +1.2] & +0.2 [-0.4, +0.8] \\
question-bank capacity $8$ (paper $16$) & -0.3 [-1.0, +0.4] & +0.9 [+0.4, +1.5] & -0.0 [-0.7, +0.6] \\
question-bank capacity $32$ & -0.6 [-1.1, -0.1] & +0.5 [-0.2, +1.2] & +0.4 [-0.3, +1.1] \\
$\alpha=5$ (paper $20$) & +0.1 [-0.5, +0.7] & +0.6 [-0.1, +1.2] & +0.6 [-0.0, +1.1] \\
$\alpha=50$ & -0.7 [-1.2, -0.3] & -0.2 [-0.9, +0.5] & -1.0 [-1.5, -0.4] \\
recalibration ridge $3$ (paper $10$) & +0.2 [-0.3, +0.6] & +0.2 [-0.3, +0.8] & -0.3 [-0.8, +0.2] \\
recalibration ridge $30$ & +0.0 [-0.4, +0.4] & +0.7 [+0.2, +1.2] & +0.2 [-0.3, +0.6] \\
recalibration refit every $5$ sessions (paper $10$) & -0.0 [-0.5, +0.5] & +0.3 [-0.1, +0.7] & -0.0 [-0.8, +0.7] \\
refit every $20$ sessions & -0.1 [-0.5, +0.4] & -0.1 [-0.4, +0.3] & -0.2 [-0.9, +0.5] \\
\bottomrule
\end{tabular}
}
\caption{Sensitivity of \method{} to its settings, changed one at a time on the three main rosters (\numSeedsU{} seeds on the held-out profiles, \numSensSeedsC{} on Composite and Quirky; never used to choose a setting): paired change in benchmark score against \method{} with the paper's settings, with a 95\% $t$ interval over seeds; negative values mean the paper's setting is better. In \numSensBelowClosest{} of the \numSensCells{} cells the changed \method{} still has lower excess regret than the closest comparison method on that roster (significantly in \numSensSigBelowClosest{}).}
\label{tab:sensitivity}
\end{table}

%% file: generated/tab_controls.tex
\begin{table}[t]
\centering
\small
\begin{tabular}{lrrr}
\toprule
control & excess of no transfer & change with \method{} & 95\% CI \\
\midrule
simulator, null population & 864.8 & -0.3\% & [-1.4\%, +0.9\%]$^\dagger$ \\
tutoring, scrambled roster & 1453.7 & +0.9\% & [-0.5\%, +2.2\%] \\
\bottomrule
\end{tabular}
\caption{Negative controls: paired change in excess regret of \method{} relative to its no-transfer ablation, as a share of the ablation's excess, where no structure is shared (simulator null population, $\beta=4$, $b=20$; scrambled development roster, $b=24$; 32 seeds). Negative values mean transfer costs regret. The interval resamples seeds and users after the fact; $^\dagger$: seeds only.}
\label{tab:controls}
\end{table}

%% file: generated/tab_ablation_gate.tex
\begin{table}[t]
\centering
\small
\begin{tabular}{llrr}
\toprule
control & policy & excess regret & change vs.\ no transfer [95\% CI] \\
\midrule
simulator null ($\beta=4$, $b=20$) & \method{} & 867.0 $\pm$ 3.5 & -0.3\% [-1.4\%, +0.9\%] \\
 & without gate & 973.6 $\pm$ 4.5 & -12.6\% [-13.6\%, -11.5\%] \\
 & no population & 868.4 $\pm$ 4.4 & -0.4\% [-1.4\%, +0.6\%] \\
\midrule
scrambled roster ($b=24$) & \method{} & 1441.2 $\pm$ 4.6 & +0.9\% [+0.0\%, +1.7\%] \\
 & without gate & 1463.5 $\pm$ 3.3 & -0.7\% [-1.5\%, +0.1\%] \\
 & no population & 1436.3 $\pm$ 3.5 & +1.2\% [+0.4\%, +2.0\%] \\
\bottomrule
\end{tabular}
\caption{The evidence gate on the two negative controls (32 seeds). Last column: paired gain over no transfer as a share of its excess regret, with a 95\% $t$ interval over seeds; negative values mean the policy costs regret where there is nothing to transfer.}
\label{tab:ablation-gate}
\end{table}

%% file: generated/tab_ablation_library.tex
\begin{table}[t]
\centering
\small
\begin{tabular}{lr}
\toprule
pair rule (profiles' tables known) & excess regret \\
\midrule
greedy & 961.8 $\pm$ 2.8 \\
one-step lookahead & 954.6 $\pm$ 2.9 \\
latent-bandit Thompson sampling & 1197.6 $\pm$ 3.8 \\
latent-bandit UCB & 1114.6 $\pm$ 1.6 \\
\bottomrule
\end{tabular}
\caption{Pair rules given the development profiles' choice probabilities and utilities (a diagnostic, not an admissible policy), development roster, 32 seeds.}
\label{tab:ablation-library}
\end{table}